\documentclass[10pt,twocolumn,letterpaper]{article}

\usepackage{iccv}
\usepackage{times}
\usepackage{epsfig}
\usepackage{graphicx}
\usepackage{amsmath}
\usepackage{amssymb}

\makeatletter
\@namedef{ver@everyshi.sty}{}
\makeatother
\usepackage{tikz}
\usetikzlibrary{shapes}
\usetikzlibrary{automata,topaths}
\usetikzlibrary{calc}

\usepackage{wrapfig}

\usepackage{caption}
\usepackage{subcaption}

\usepackage{floatrow}

\usepackage{ragged2e}

\usepackage{afterpage}

\usepackage{rotating}
\usepackage{makecell}
\usepackage{enumitem}

\usepackage{mathtools}

\usepackage{snapshot}

\usepackage[toc,page]{appendix}

\usepackage[
backend=biber,
style=ieee,
citestyle=ieee,
]{biblatex}
\AtBeginBibliography{\footnotesize}

\newcommand{\imageone}[1]{}
\newcommand{\imagetwo}[1]{}

\makeatletter
\newcommand{\explainedstep}[1]{%
	\par
	\textbf{#1}\enspace\ignorespaces
}
\makeatother

\usepackage{microtype}

\RequirePackage[none]{hyphenat}
\def\theLetterSpace{-0.5pt}
\def\extraWordSpace{-0.5pt}
\newcommand\spaceout[2][\theLetterSpace]{%
  \def\LocalLetterSpace{#1}\expandafter\spaceouthelpA#2 \relax\relax}
\def\spaceouthelpA#1 #2\relax{%
  \spaceouthelpB#1\relax\relax%
  \ifx\relax#2\else\kern\extraWordSpace\ \kern\LocalLetterSpace\spaceouthelpA#2\relax\fi
}
\def\spaceouthelpB#1#2\relax{%
  #1%
  \ifx\relax#2\else
    \kern\LocalLetterSpace\spaceouthelpB#2\relax%
  \fi
}

\newif\ifdraft
\draftfalse

\ifdraft
\newcommand{\fpc}[1]{{\color{cyan}\textbf{FP:} #1}}
\newcommand{\dcc}[1]{{\color{red}\textbf{DC:} #1}}
\newcommand{\odc}[1]{{\color{orange}\textbf{OD:} #1}}
\newcommand{\abc}[1]{{\color{magenta}\textbf{AB:} #1}}

\else
\newcommand{\fpc}[1]{}
\newcommand{\dcc}[1]{}
\newcommand{\odc}[1]{}
\newcommand{\abc}[1]{}

\fi

\usepackage[breaklinks=true,bookmarks=false]{hyperref}

\iccvfinalcopy %

\def\iccvPaperID{4961} %

\begin{document}

\title{Pix2Vex: Image-to-Geometry Reconstruction\\ using a Smooth Differentiable Renderer}

\newcommand{\negkern}{\kern-.0em}
\newcommand{\negkernOut}{\kern-.75em}
\author{
    \negkernOut Felix Petersen\negkern \\
    \negkernOut University of Konstanz\negkern \\
    \negkernOut {\tt\small felix.petersen@uni.kn}\negkern 
    \and
    \negkern Amit H. Bermano\negkern \\
    \negkern Tel Aviv University\negkern \\
    \negkernOut {\tt\small \spaceout{amberman@tauex.tau.ac.il}}\negkernOut
    \and
    \negkern Oliver Deussen\negkern \\
    \negkern University of Konstanz\negkern \\
    \negkern {\tt\small oliver.deussen@uni.kn}\negkern
    \and
    \negkern Daniel Cohen-Or\negkernOut \\
    \negkern Tel Aviv University\negkernOut \\
    \negkern {\tt\small dcor@tau.ac.il}\negkernOut 
}

\maketitle
\begin{abstract}

The long-coveted task of reconstructing 3D geometry from images is still a standing problem. 
In this paper, we build on the power of neural networks and introduce \texttt{Pix2Vex}, a network trained to convert camera-captured images into 3D geometry.
We present a novel differentiable renderer ($DR$) as a forward validation means during training. 
Our key insight is that $DR$s produce images of a particular appearance, different from typical input images. 
Hence, we propose adding an image-to-image translation component, converting between these rendering styles. 
This translation closes the training loop, while allowing to use minimal supervision only, without needing any 3D model as ground truth.
Unlike state-of-the-art methods, our $DR$ is $C^\infty$ smooth and thus does not display any discontinuities at occlusions or dis-occlusions. 
Through our novel training scheme, our network can train on different types of images, where previous work can typically only train on images of a similar appearance to those rendered by a $DR$.

\end{abstract}

\begin{figure}[htb]
\centering
	\begin{tabular}{c}
	    \subcaptionbox{direct approach \label{fig:naiive}}{\includegraphics[width=.8\linewidth]{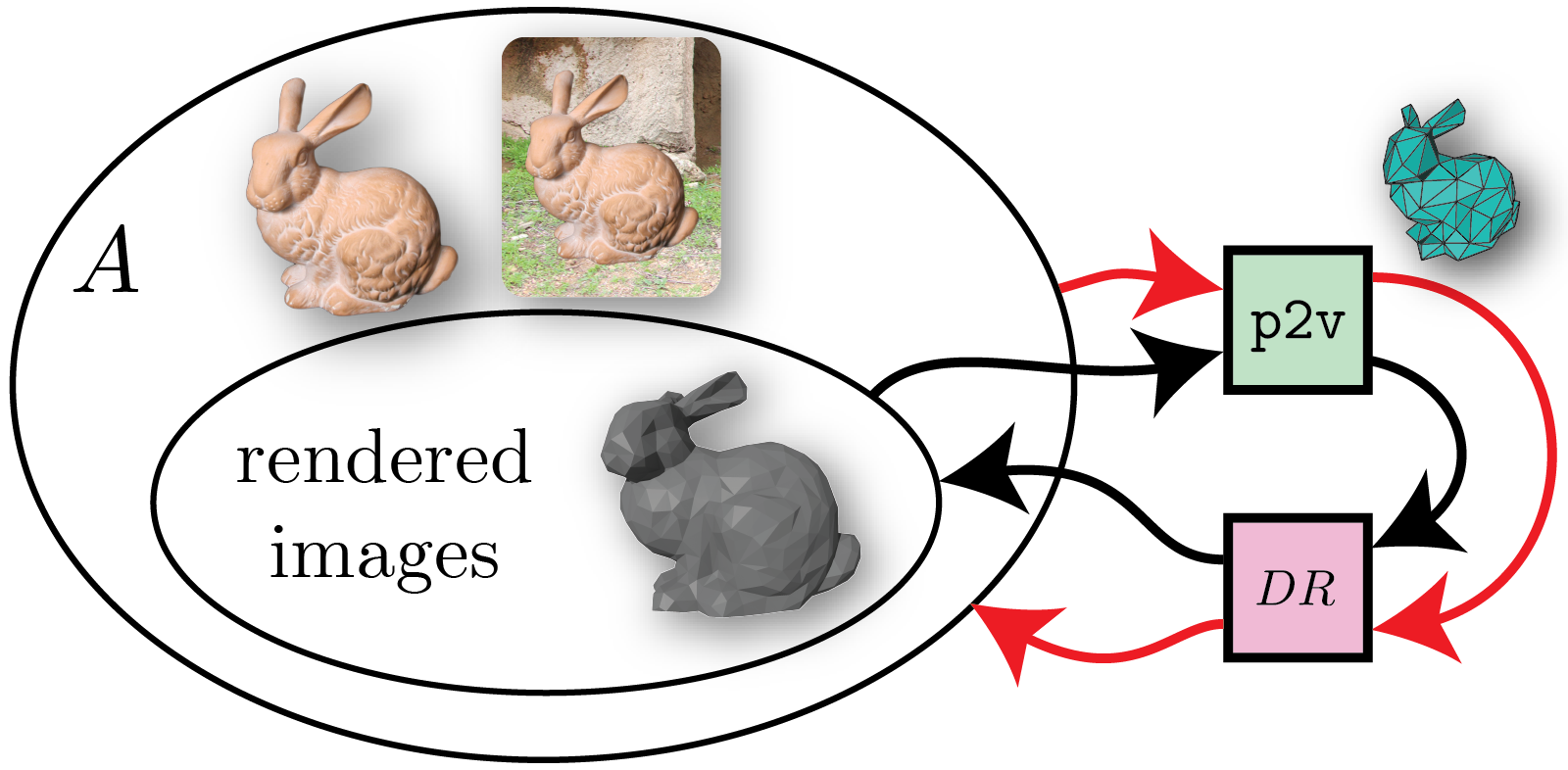}}
		\\~\\
		\subcaptionbox{novel indirect approach \label{subfig:novelvenn}}{\includegraphics[width=.8\linewidth]{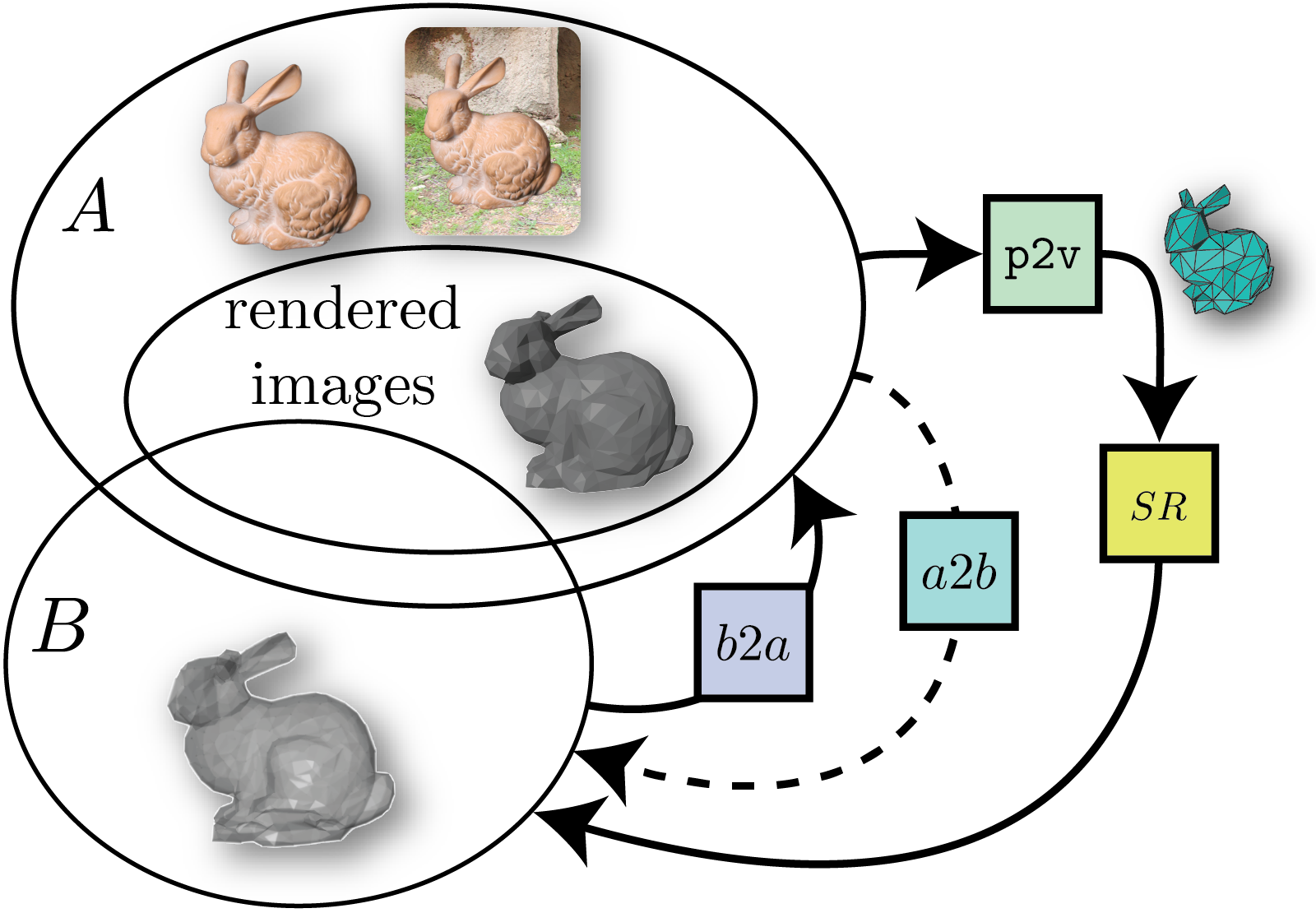}}
	\end{tabular}
\label{subfig:novelvenn}
\caption{
	Traditional reconstruction approaches employing Differentiable Renderers ($DR$s) rely on differentially rendered images alone ((a), black path), and would fail when used with generally rendered images ((a), red path). We propose a novel \textit{indirect} approach, in which domain translation components help in closing the training loop (b).
	}
\end{figure}
\section{Introduction}

Reconstructing a 3D model from arbitrary images has been considered one of the big challenges
of computer vision since the inception of the field. This ambitious task encapsulates many challenges, including separation of texture and shading effects, lighting estimation, and inference of occluded parts. Hence, different variants of the problem, which take advantage of different assumptions, have been tackled over the years, employing a myriad of approaches. 

Perhaps the most interesting and promising approaches are those utilizing learning-based methods \cite{Che2018, Henderson2018, Wang2018, Kanazawa2018}.
These can potentially exploit previously seen examples to overcome missing information such as partial visibility, unknown lighting conditions, and others.
To date, most of these methods either rely on extensive manual annotation of shape \cite{Wang2018} or key-point correspondences between images \cite{Kanazawa2018}.
Other, less supervised methods \cite{Che2018, Henderson2018} are typically restricted to settings of controlled materials, colors, lighting, and/or precise annotation of object orientation.

In this paper, we validate a predicted 3D reconstruction using a renderer, thus closing the image-to-geometry (\texttt{pix2vex}) and geometry-to-image (renderer) training loop.
In our setting, we feed the system with grayscale images of an object from one or more directions (usually four, see Section \ref{sec:res}), and predict its corresponding 3D geometry. In contrast to popular approaches, we provide a rough estimate of the angular difference between the aforementioned images and leave all other parameters to be deduced by the system. This means that our method does not require lighting and texture information, nor explicit 3D supervision. 

This setting poses a threefold challenge. In the following, we specify these challenges, and our respective proposed solutions, which constitute the main contributions of this paper:

First, the success of such an approach depends greatly on the ability to generate meaningful gradients throughout the validation process.
State-of-the-art differentiable renderers ($DR$s), however, besides being bounded by complexity and output quality, are only locally differentiable with respect to the geometry.
The differentiability (or smoothness) breaks in cases of occlusion and dis-occlusion cases (see Section~\ref{sec:smooth} for details).
Hence, we present a novel $C^\infty$ \textbf{smooth differentiable renderer}, $SR$, based on soft blending schemes of nearby triangles.
This unique property is the backbone of our approach, enabling learning with minimal supervision. 

Secondly, we present the image-to-geometry neural network, \textbf{\texttt{pix2vex}}, that is trained to predict the 3D coordinates of a given mesh topology.
The network seeks to find vertex positions that produce renderings that match the input images, while maintaining shape smoothness with minimal supervision.
A direct solution would use an auto-encoder--like network, as proposed by Che \textit{et al.} \cite{Che2018}.
This approach, however, restricts the network training to images that have a similar appearance to the output of the employed forward renderer (see Figure \ref{fig:naiive}).
In order to extend the space of images that can be used for training, we present the following insight.

Due to their differential nature, $DR$s (and $SR$s) produce images which are somewhat different compared to traditional renderers.
To bridge this gap, we employ an image-to-image translator, $b2a$, that converts images from our $SR$'s output domain, $B$, to images in the input one, $A$, and thereby closes the training loop, as depicted in Fig. \ref{subfig:novelvenn}.
Furthermore, it turns out that this three-component-chain (\texttt{pix2vex}, $SR$, $b2a$) is extremely hard to train without supervision.
Thus, we constrain the learning by employing another image-to-image translator, $a2b$, that translates from the input domain, $A$, to the output domain of our $SR$, $B$.
To train these components (\texttt{p2v}, $a2b$, $b2a$), we introduce a \textbf{Reconstructive Adversarial Network (RAN)}.
As can be seen in Section \ref{sec:ran}, training a RAN has strong similarities to training Generative Adversarial Networks (GANs) and especially Conditional Adversarial Networks.
Like GANs, the RAN has a discriminator $D$, which plays an adversarial role.
However, $D$ in our case is trained to tell whether a pair of images from $A$ and $B$ comply (i.e., have similar content and respective styles), rather than to indicate whether the given image belongs to the target distribution or not.

\begin{figure*}[t]
	\centering
	\includegraphics[width=.6\linewidth]{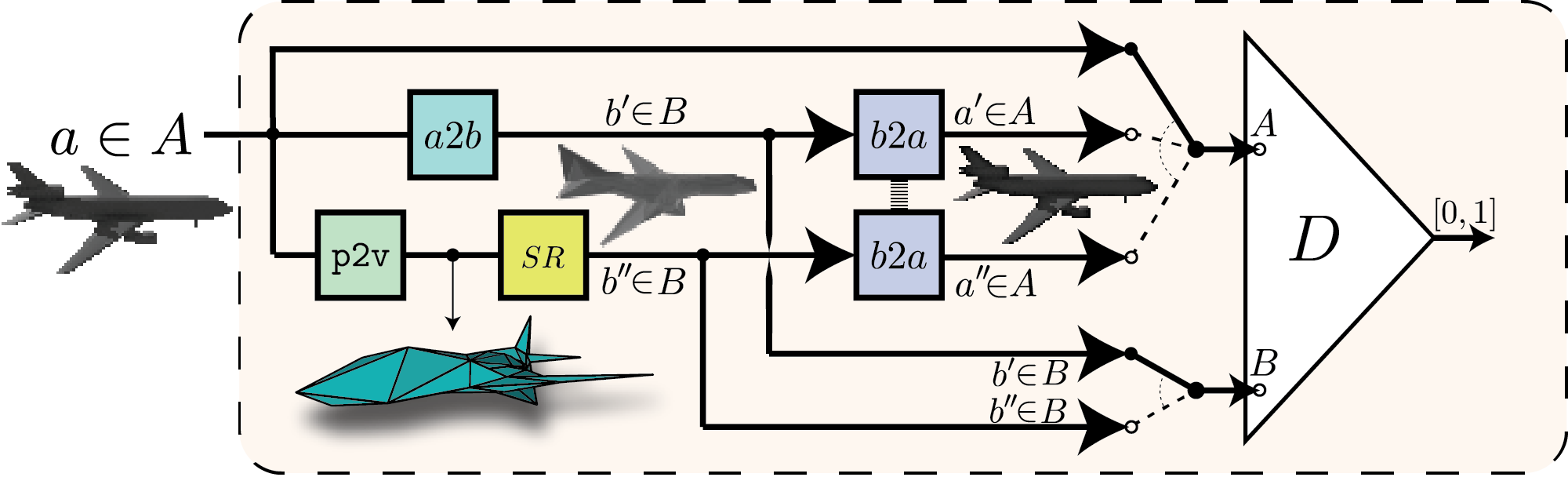}
	\captionsetup{width=1.\textwidth}
    \caption{
	RAN System overview. Our image-to-geometry reconstructor (\texttt{p2v}) receives a set of images from the input domain $A$ (e.g., grayscale images rendered by an off-the-shelf renderer) and predicts vertex positions that match. 
	The reconstruction is validated through our novel $C^\infty$ smooth renderer ($SR$).
	The latter produces images in a different domain, $B$, which are translated back to the input domain $A$ for training purposes ($b2a$). 
	Unlike in traditional $DR$-based GAN systems, the purpose of our discriminator $D$ is to indicate whether the two inputs match in content.
	The whole network is trained via five different data paths, including two which require another image domain translator, $a2b$, in a novel training scheme. 
	\vspace{1em}
	}
	\label{fig:rancomb}
\end{figure*}

These contributions constitute our novel architecture, consisting of the five elements depicted in Figure \ref{fig:rancomb}.
The smooth renderer is explained in more detail in Section \ref{sec:smooth}.
The other components are simultaneously trained in a non-trivial manner through different data paths, as portrayed in Section \ref{sec:ran}.
Once the prediction network \texttt{pix2vex} (Section \ref{sub:reconstructor}) has been trained, it is able to predict the 3D geometry of an object from one or more of its images.
We show the results of this method in Section \ref{sec:res}.

\section{Related Work}\label{sec:rel}

Reconstructing 3D models from single or multiple images has been approached with different techniques. 
A number of recent works \cite{Fan2016, Achlioptas2017, Choy2016, Wang2018, Jiang2018, Kurenkov2018} have developed 3D reconstruction techniques based on supervision.
In the following, we focus on methods that employ differentiable renderers.

The most common and efficient representation for 3D models is irregular polygonal meshes.
Other representations, like voxel grids
\cite{Gadelha2016, Henzler2018, Rezende2016, Yan2016, Tulsiani2017, Gwak2017, Zhu2018, Wu2017},
or points clouds
\cite{L2018, Jiang2018}
are typically of lower resolution and do not scale well. 
Moreover, smoothly shading their discrete surfaces is difficult, and typically restricts to rendering only silhouettes \cite{Gadelha2016, Yan2016, Gwak2017, Jiang2018, Tulsiani2017, L2018}.
Working directly on meshes allows higher visual and computational quality while keeping memory footprint to a minimum.

Numerous works
\cite{Kato2017, Loper2014, Li2018, Bangaru2019, Henderson2018, Che2018, Liu2018, Delaunoy2011, Ramamoorthi2001, Meka2018, Athalye2017, Richardson2017, Palazzi2019, Liu2019}
have presented differentiable renderers for 3D meshes.
These renderers are differentiable with respect to lightning,
geometry
\cite{Che2018, Li2018, Loper2014, Kato2017, Liu2018, Delaunoy2011, Richardson2017, Bangaru2019, Liu2019},
material
\cite{Ramamoorthi2001, Meka2018, Athalye2017, Bangaru2019},
and/or texture
\cite{Ramamoorthi2001, Athalye2017, Liu2018}.
As discussed in Section \ref{sec:smooth}, however, these renderers are not differentiable on the entire domain (with respect to geometry, which is the relevant aspect when considering shape reconstruction).
It is also possible to estimate a differentiable renderer with a Render-and-Compare--loss, which employs finite differences \cite{Kundu2018}.
As can be expected, such an approach comes at the cost of computational effort and accuracy, compared to analytic gradients.

Rather than predicting, some works perform a mesh optimization process on a single 3D object
\cite{Delaunoy2011, Li2018, Bangaru2019, Loper2014, Kato2017, Liu2018}.
In these works, an initial mesh is optimized by iteratively changing its geometry, guided by back-propagated gradients.
The main issue with such an approach is computational cost: this computationally expensive optimization process has to be repeated for every reconstruction.
Prediction, on the other hand, is orders of magnitude faster~\cite{Che2018, Henderson2018, Richardson2017, Kato2017, Liu2019}.
For prediction, a proposed approach is to use auto-encoder--like networks to train a 3D mesh predicting encoder.
If an object, and not just its silhouette, is differentially rendered, the auto-encoder--like network has an encoder-renderer structure predicting the 3D model in the latent space 
\cite{Che2018, Henderson2018, Richardson2017}.
These encoder-render methods are typically restricted to reconstructing from images that match images produced by the employed differentiable renderer in appearance and style.
Other works that render only the silhouette of an object
\cite{Kato2017, Gadelha2016, Yan2016, Liu2019}
are only trained towards predicting proper ground truth silhouettes, while ignoring normals, and thus lighting.

Our work, in particular, the RAN, is closely related to Generative Adversarial Networks (GANs) 
\cite{Goodfellow2014},
which allow generating new samples from a distribution, e.g., for generating voxel grids for a given class of objects
\cite{Henzler2018, Gadelha2016, Gwak2017, Zhu2018, Rezende2016}.
These GANs can also be employed to assist 3D reconstruction by supervising the appearance of the reconstructed models
\cite{Gadelha2016, Gwak2017, Rezende2016}.

Finally, there are neural networks that are trained under supervision to render silhouettes from voxel grids
\cite{Nguyen-Phuoc2018, Rezende2016}
or shaded images from a latent space
\cite{Eslami2018}. 
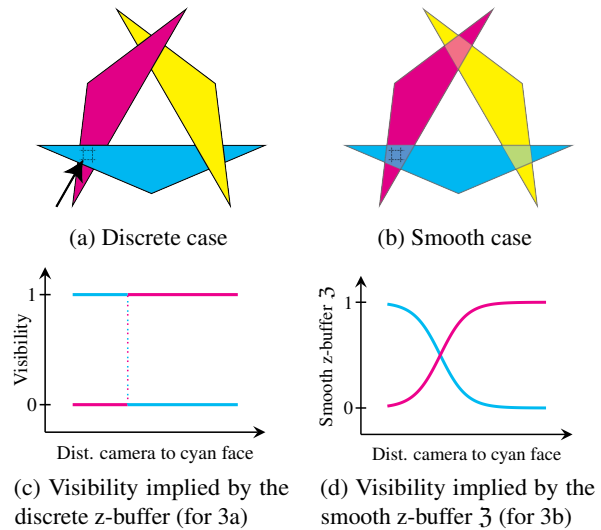
\begin{figure}[b]
	\centering
	\usetikzlibrary{arrows}
	\def\layersep{1}
	\def\stepoffset{4}
	\begin{tabular}{cc}
		\subcaptionbox{Discrete case \label{subfig:discrete}}{\hspace*{-.5em}\includegraphics[width=.37\linewidth]{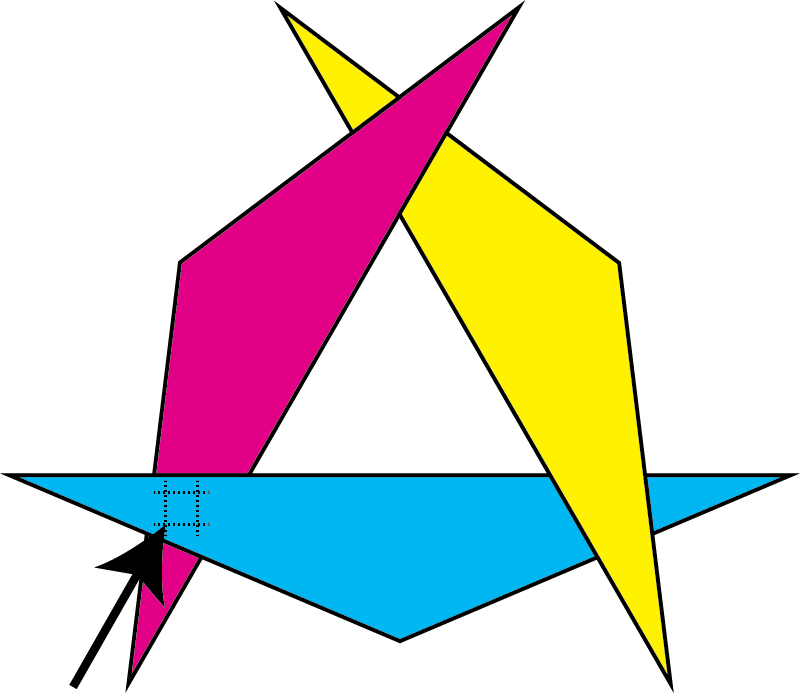}\hspace*{-.5em}} &
		\subcaptionbox{Smooth case \label{subfig:smooth}}{\hspace*{-.5em}\includegraphics[width=.37\linewidth]{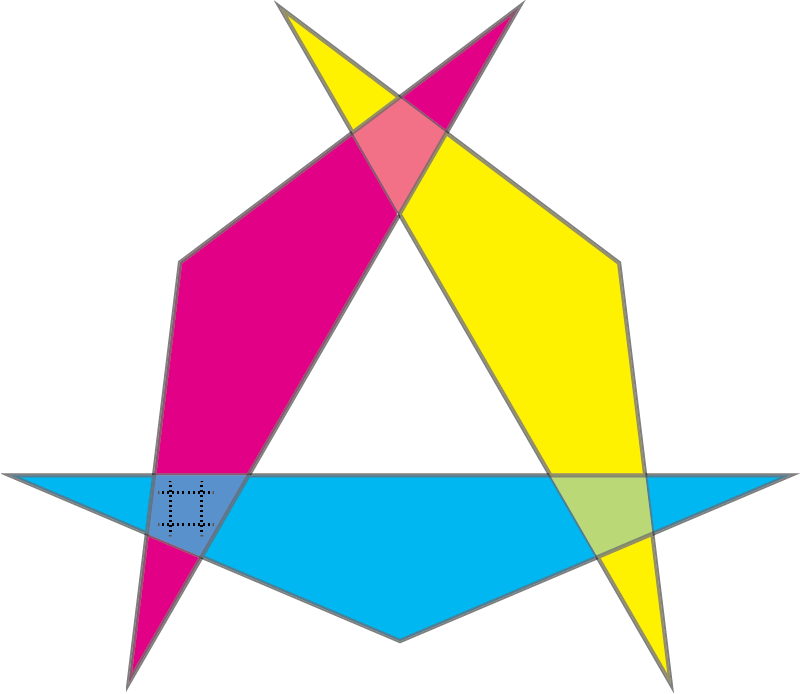}\hspace*{-.5em}} \\
		\subcaptionbox{Visibility implied by the discrete z-buffer (for \ref{subfig:discrete}) \label{subfig:discreteplot}}{\hspace*{-1.em}\resizebox{.48\linewidth}{!}{%
\begin{tikzpicture}[scale=0.65]

\tikzstyle{arrow} = [->,>=stealth]

	\node[](n21) {
		\begin{tikzpicture}[node distance=\layersep,scale=.5]

		\draw[color=magenta, thick] plot[
		mark=none,
		samples=100,
		domain=-1:0,
		] ({\x},{0});;
		\draw[color=magenta, thick] plot[
		mark=none,
		samples=100,
		domain=0:2,
		] ({\x},{2});;
		\draw[color=cyan!80, thick] plot[
		mark=none,
		samples=100,
		domain=-1:0,
		] ({\x},{2});;
		\draw[color=cyan!80, thick] plot[
		mark=none,
		samples=100,
		domain=0:2,
		] ({\x},{0});;
		
		\draw[dotted, cyan!80] (0, 0) -- (0, 2);
		\draw[dotted, magenta] (0, 0.1) -- (0, 2);
		
		\draw[arrow, draw=black] (-1.5, -.5) -- node[anchor=south, yshift=0cm, xshift=0cm, rotate=90] {\tiny Visibility} (-1.5, 2.5) ;
		\draw[arrow, draw=black] (-1.5, -.5) -- node[anchor=north, yshift=0cm, xshift=0cm] {\tiny Dist. camera to cyan face} (2.5, -.5) ;
		
		\draw[draw=black] (-1.6, 0) -- node[anchor=east, yshift=0cm, xshift=.05cm] {\tiny 0} (-1.4, 0) ;
		\draw[draw=black] (-1.6, 2) -- node[anchor=east, yshift=0cm, xshift=.05cm] {\tiny 1} (-1.4, 2) ;

		\end{tikzpicture}
	};
	
\end{tikzpicture} %
}\vspace*{-.4cm}\hspace*{-.0em}} &
		\subcaptionbox{Visibility implied by the smooth z-buffer $\mathfrak{Z}$ (for \ref{subfig:smooth}) \label{subfig:smoothplot}}{\hspace*{-1.em}\resizebox{.48\linewidth}{!}{%
\begin{tikzpicture}[scale=0.65]

\tikzstyle{arrow} = [->,>=stealth]
	
	\node[](n31) {
		\begin{tikzpicture}[node distance=\layersep,scale=.5]

		\draw[color=cyan!80, thick] plot[
		mark=none,
		samples=100,
		domain=-1:2,
		] ({\x},{2/(1+exp(4*\x))});;
		\draw[color=magenta, thick] plot[
		mark=none,
		samples=100,
		domain=-1:2,
		] ({\x},{2/(1+exp(-4*\x))});;
		
		\draw[arrow, draw=black] (-1.5, -.5) -- node[anchor=south, yshift=0cm, xshift=-.125cm, rotate=90] {\tiny Smooth z-buffer $\mathfrak{Z}$} (-1.5, 2.5) ;
		\draw[arrow, draw=black] (-1.5, -.5) -- node[anchor=north, yshift=0cm, xshift=0cm] {\tiny Dist. camera to cyan face} (2.5, -.5) ;
		
		\draw[draw=black] (-1.6, 0) -- node[anchor=east, yshift=0cm, xshift=.05cm] {\tiny 0} (-1.4, 0) ;
		\draw[draw=black] (-1.6, 2) -- node[anchor=east, yshift=0cm, xshift=.05cm] {\tiny 1} (-1.4, 2) ;
		
		\end{tikzpicture}
	};

\end{tikzpicture} %
}\vspace*{-.4cm}\hspace*{-.0em}}
	\end{tabular}
	\caption{
	    Visualization of the smooth depth buffer and occlusion:
		\ref{subfig:discrete} shows three triangles rendered in a standard way, in \ref{subfig:smooth} the same triangles are rendered smoothly.
		While in the discrete case a small change in depth can result in a sudden change of color (\ref{subfig:discreteplot}), our smooth depth-oriented rendering (\ref{subfig:smoothplot}) avoids that and therefore is differentiable everywhere.
	}
	\label{fig:occlusion}
\end{figure}

\begin{figure*}[t]
	{\begin{tabular}{ccccc}
	        \hspace{-1.5em}
			\subcaptionbox{low $s$, low $o$\label{subfig:lslo}}{\includegraphics[width=.2\linewidth]{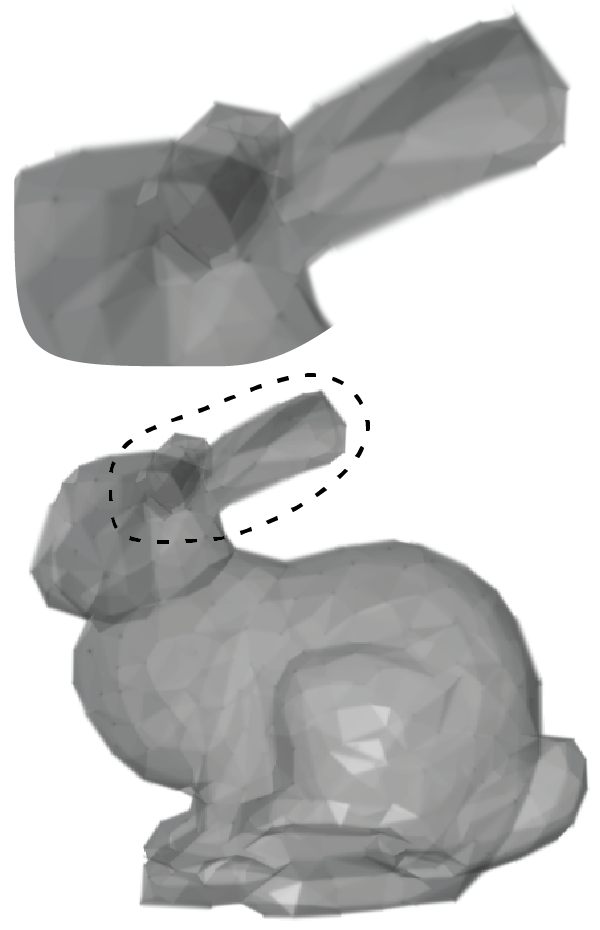}} \hspace{-1.5em} & 
			\subcaptionbox{high $s$, low $o$\label{subfig:hslo}}{\includegraphics[width=.2\linewidth]{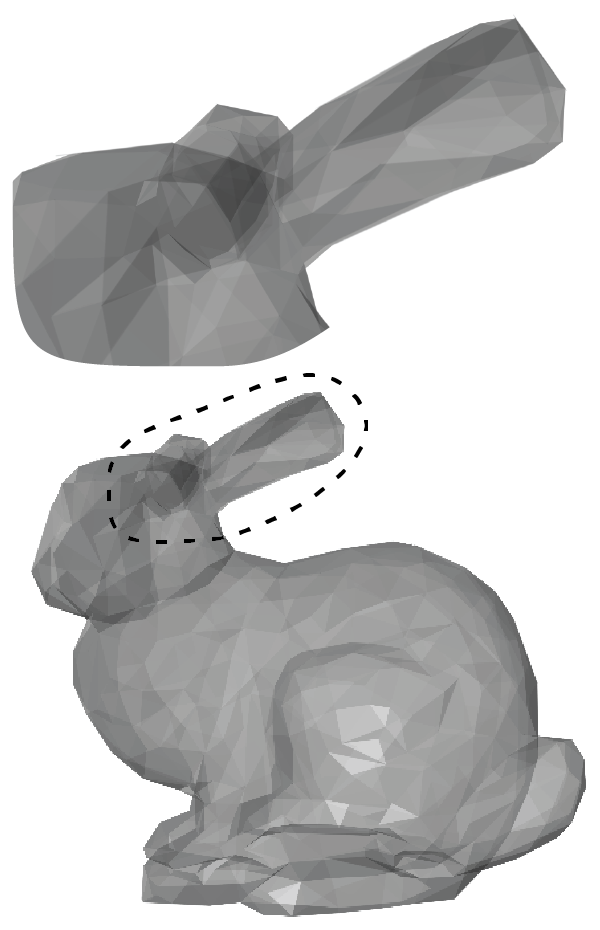}} \hspace{-1.5em} & 
			\subcaptionbox{low $s$, high $o$\label{subfig:lsho}}{\includegraphics[width=.2\linewidth]{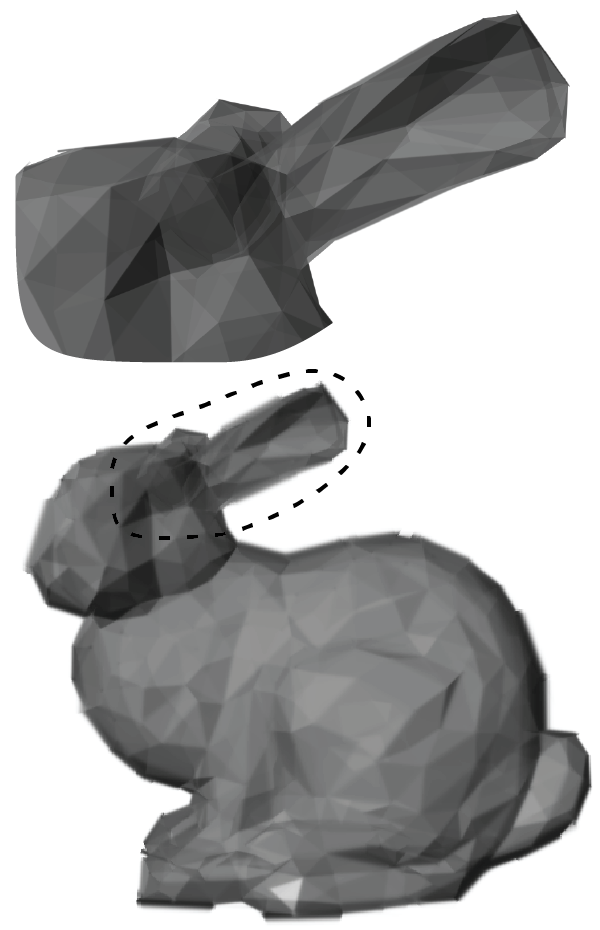}} \hspace{-1.5em} & 
			\subcaptionbox{high $s$, high $o$\label{subfig:hsho}}{\includegraphics[width=.2\linewidth]{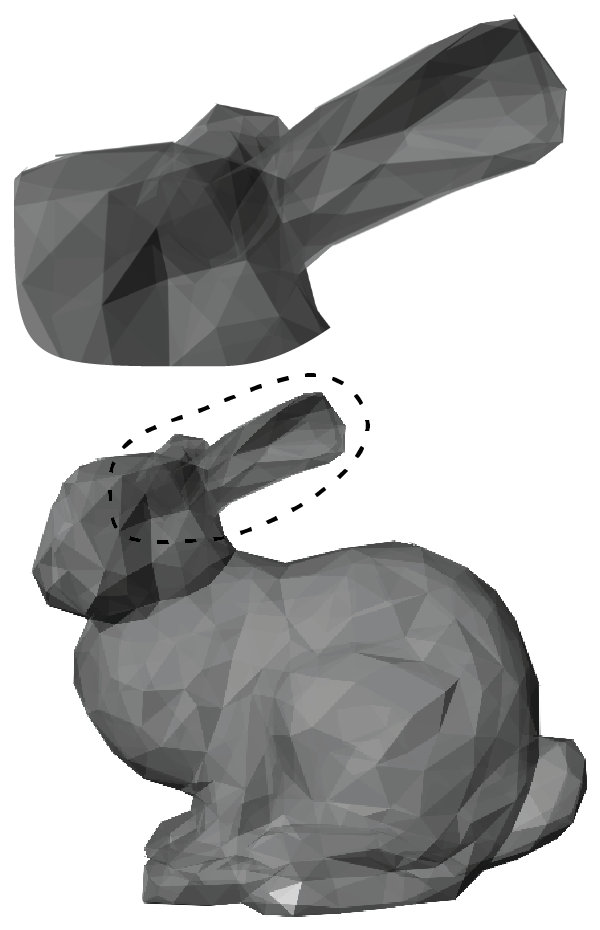}} \hspace{-1.5em} & 
			\subcaptionbox{Blender result}{\includegraphics[width=.2\linewidth]{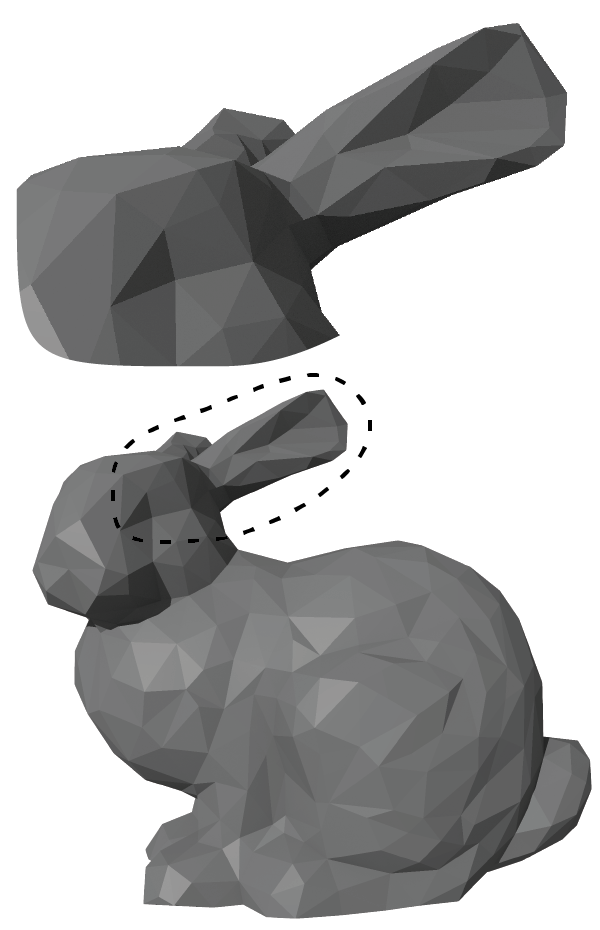}}  
			\hspace{-1.5em}
		\end{tabular}\\} 
	\captionsetup{width=.95\linewidth}
	\caption{
	Stanford bunny rendered by the smooth renderer using different edge smoothnesses ($s$) and opacities ($o$): 
	In \ref{subfig:lslo} and \ref{subfig:hslo}, the low opacity $o$ causes, e.g., one of the ears to be still visible through the head of the bunny (for usage of $o$ see section \ref{sub:smoothzbuffer}).
	In \ref{subfig:lslo} and \ref{subfig:lsho}, the low edge steepness causes smoother edges (for usage of $s$ see section \ref{sub:visibilitytensor}). 
	On the right: The Stanford bunny rendered by Blender (e).
	}
	\label{fig:renderings}
\end{figure*} 
\section{$C^\infty$ Smooth Renderer}\label{sec:smooth}

In this section, we present our $C^\infty$ Smooth Renderer that avoids any discontinuities at occlusions or dis-occlusions.
Having this property, the renderer's back-propagated gradients can be properly used to modify the 3D model.
This is critical for integrating the renderer %
into a neural network.
The typical discontinuity problem occurs during triangle rasterization, where the visibility of a triangle, due to occlusion or dis-occlusion, causes an abrupt change in the image.
For example, if during the optimization process, the backside of a predicted object self-intersects its front, traditional differentiable renderers are not able to provide a reasonable gradient towards reversing such an erroneous self-intersection since they cannot differentiate with respect to occlusion.
To overcome this problem, our approach offers a soft blending scheme, that is continuous even through such intersections.
As in the general rendering approach, first, we apply view transformations on all triangles to bring them from object space into perspective projection space coordinates.
This process is generally already fully differentiable.

Consecutively, one needs rasterization to correlate triangles to pixels.
General rasterization consists of two steps, for each pixel one needs to collect all the triangles that cover that pixel, and then employ a z-buffer to determine which of them is visible in the pixel. 

Instead of collecting all triangles that fit the xy-coordinates of a given pixel, we determine a probability value of whether a triangle fits a pixel for each triangle and pixel.
This constitutes the visibility tensor $V$ as shall be described in section \ref{sub:visibilitytensor}.

Our key idea is to use a visibility test that enables reasoning beyond occlusion, using only smooth functions to avoid abrupt changes. 
Rather than taking a discrete decision of which triangle is the closest and thus visible, we softly blend their visibility, which goes along with an idea from stabilizing non-photorealistic rendering results~\cite{Luft2006}.
By using a SoftMin-based function, we determine the closest and thus most visible face.
But using the simple SoftMin of the z-positions in camera space would result in only the single closest triangle being most visible.
Thus, we need to incorporate the visibility tensor $V$ that tells us which triangles cover a given pixel.
Instead, we weight the SoftMin with the visibility tensor $V$ by introducing the weighted SoftMin ($\mathfrak{w}$SoftMin).
Taking the $\mathfrak{w}$SoftMin of the z-positions in camera space, constitutes smooth z-buffer as shall be described in greater detail in section \ref{sub:smoothzbuffer}.

This smooth z-buffer leads to our $C^\infty$ Smooth Renderer, where the z-positions of triangles is differentiable with respect occlusions.
In previous differentiable renderers, only the xy-coordinates were locally differentiable with respect to occlusion.

Let us assume to have three triangles (see Fig. \ref{subfig:discrete}), where we want to examine the behavior of the bottom left pixel (marked with \#) with respect to the z-position of the cyan face.
During the process of optimizing the geometry, triangles might change their order with regard to the depth and abrupt color changes might appear.
As shown in Figure \ref{subfig:discreteplot}, the color value of the pixel (implied by the rasterization of the triangles) is constant except for one single point.
At this point of intersection, the rasterization is discontinuous; at all other points, the derivative with respect to the z-position is 0.
Employing the smooth rasterization and smooth z-buffer as in Figure \ref{subfig:smoothplot}, the visibility of a pixel is never absolute, but rather a soft blend.
Thus, it is differentiable, and optimizations can be solved with simple gradient descent.

Finally, we need to compute the color values of the triangles.
For that, we use a lightning model composed of Blinn-Phong, diffuse, and ambient shading.
We restrict the color to grayscale since we do not reconstruct the color in the RAN.
Since the function of color is already differentiable we can directly use it.

Figure \ref{fig:renderings} shows a comparison between our smooth renderings and a Blender rendering of the Stanford bunny.

\begin{figure*}[t]
	\centering
    \resizebox{\linewidth}{!}{
	\begin{tabular}{ccccc}
		\subcaptionbox{check whether point is in triangle \label{subfig:in-tri}}{
		    \includegraphics[width=.27\linewidth]{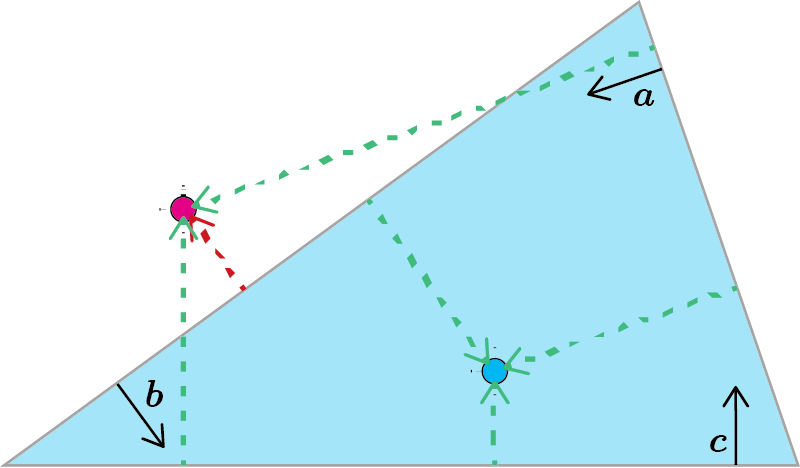}
		    } &
		\subcaptionbox{sigmoid wrt. edge $a$ \label{subfig:tri-sig-a}}{
    		\includegraphics[width=.16\linewidth]{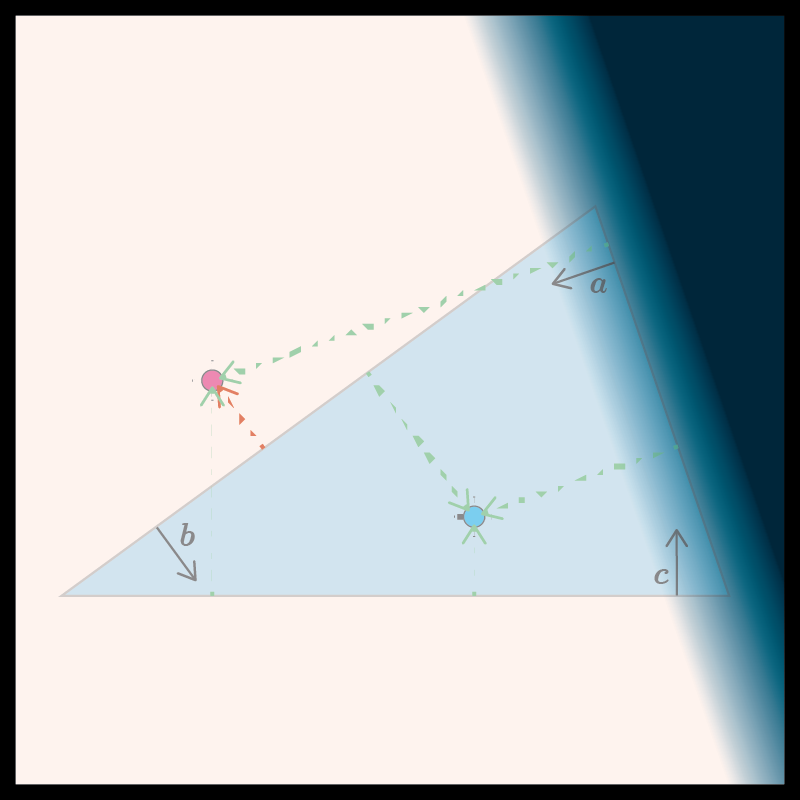} 
		    } &
		\subcaptionbox{sigmoid wrt. edge $b$ \label{subfig:tri-sig-b}}{
    		\includegraphics[width=.16\linewidth]{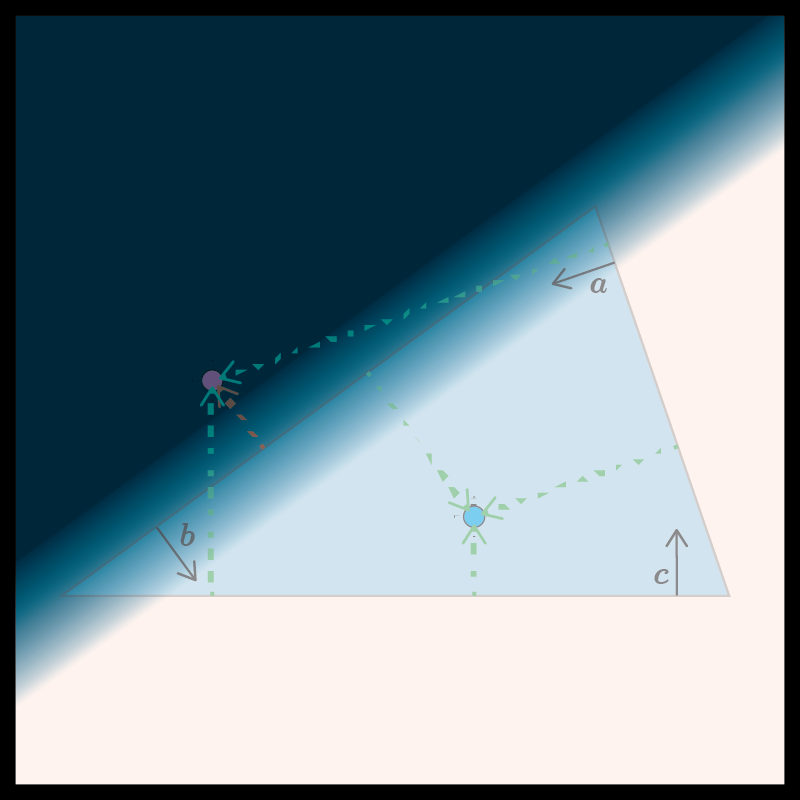} 
		    } &
		\subcaptionbox{sigmoid wrt. edge $c$ \label{subfig:tri-sig-c}}{
    		\includegraphics[width=.16\linewidth]{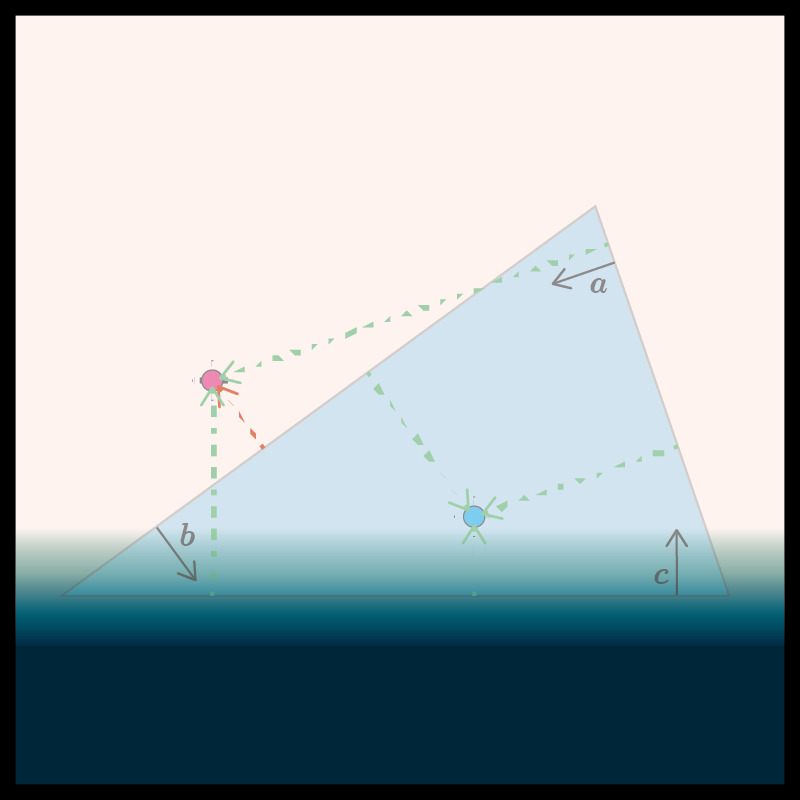} 
		    } &
		\subcaptionbox{product of sigmoids 
    		\label{subfig:tri-sig-comb}}{
    		\includegraphics[width=.16\linewidth]{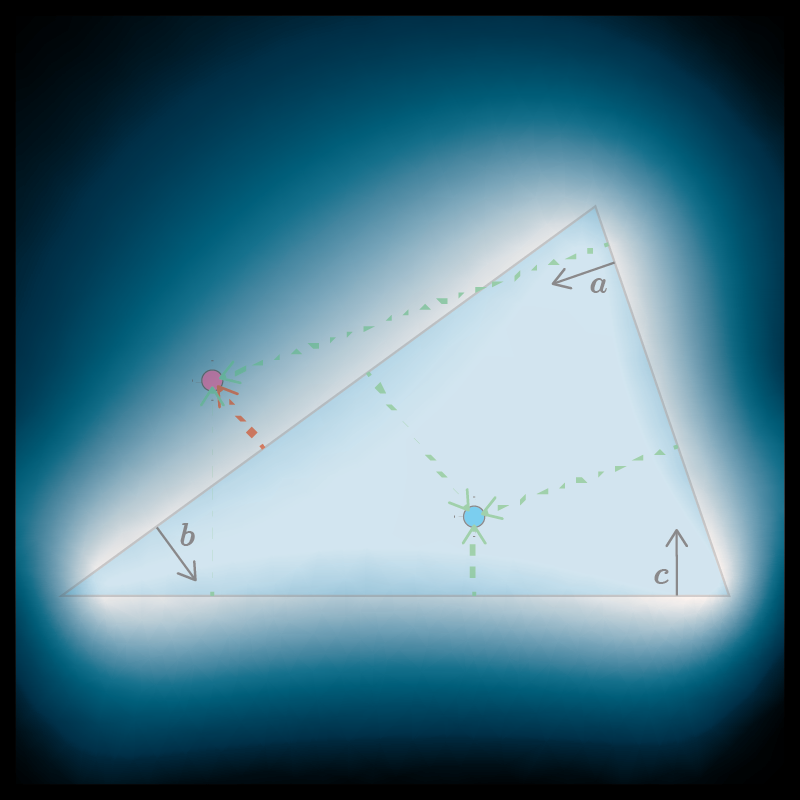}
		    }
	\end{tabular}
	}
	\captionsetup{width=.95\linewidth}
	{\caption{
			Visualization of the smooth rasterization.
			While the magenta point lies outside the triangle, the cyan point lies inside the triangle; this can be determined by measuring on which side of the edges a point lies.
			In subfigure \ref{subfig:tri-sig-a}--\ref{subfig:tri-sig-c} it is smoothly determined which parts of the image lie inside and which parts lie outside the triangle with respect to the edges $a$--$c$.
			This is combined by multiplication (visibility tensor $V$) in subfigure \ref{subfig:tri-sig-comb}.
		}
		\label{fig:intriangle}}
\end{figure*}

\subsection{Smooth Z-buffer $\mathfrak{Z}$}\label{sub:smoothzbuffer}

Our rasterization step is similar to the z-buffer algorithm, but instead of a displaying the single closet triangle and its z-distance in each pixel, we display a blend of triangles that project to the pixel.

We define the Smooth z-buffer $\mathfrak{Z}$ for pixel $(p_x, p_y)$, triangle $T$, and opacity $o$ as follows:
\begin{align*}
    \mathfrak{Z}(p_x, p_y, T) :=\ \\ 
    \mathfrak{w}\mathrm{SoftMin}( & o \cdot \text{z-dist}(\mathrm{camera}, T), 
    & V(p_x, p_y,\ \cdot\ ))
\end{align*}

We define the weighted SoftMin (analogue to SoftMin/SoftMax) as:
$\mathfrak{w}\mathrm{SoftMin}(\mathbf{x}, w) := \mathfrak{w}\mathrm{SoftMax}(-\mathbf{x}, w)$
where the weighted SoftMax is defined as: 
\begin{align*}
    \mathfrak{w}\mathrm{SoftMax}_i(\mathbf{x}, w)
    :&= \frac{\exp ({\mathbf{x}_i}) \cdot w_i}{\sum_{i=0}^{\|w\|-1} \exp ({\mathbf{x}_i}) \cdot w_i} \\
    &= \text{SoftMax}_i({\mathbf{x}_i} + \log w_i)
\end{align*}

Thus, for a pixel, the closest triangle is represented with high visibility, while triangles further away have weaker visibility. 
The visibility tensor $V$, as shall be defined in Section \ref{sub:visibilitytensor}, contains the extent to which a given triangle covers a given pixel.
We use it as a weight for the $\mathfrak{w}\mathrm{SoftMin}$, which allows considering only the relevant triangles in the SoftMin operator.

The opacity $o$ is a hyper-parameter setting accelerating the strength of the SoftMin. 
See Figure \ref{fig:renderings} to see how it affects the results.

Similar to the painter's algorithm \cite{deBerg1993} we do not explicitly handle special cases of cyclical overlapping polygons that can cause depth ordering errors.
Our smooth renderer is not sensitive to these cases.
When polygons have a similar distance to the camera, their opacity will also be very similar and thus not only the front polygon but also the one behind is visible.

\subsection{Visibility tensor $V$}\label{sub:visibilitytensor}

In the general rendering approach, the discrete choice, whether a triangle covers a pixel is just a trivial check.
In the smooth approach, as shown in Figure \ref{fig:intriangle}, we determine the pixels that correspond to a triangle by checking for each pixel whether the directed distances from the pixel to each edge are all positive.
This yields the visibility tensor $V$ for triangle $T=(e_1, e_2, e_3)$ with $e_i=(v_1, v_2)$ and $v_i=(v_{i, x}, v_{i, y})$, and steepness $s$ as follows:
\begin{align*}
	V(p_x, p_y, T) := &\\
	\prod_{e=(v_1, v_2)\in T} & \sigma \left( 
	\begin{vmatrix}
	v_{x, 2}-v_{x, 1} & v_{x, 1}-p_{x} \\
	v_{y, 2}-v_{y, 1} & v_{y, 1}-p_{y}
	\end{vmatrix}
	\cdot \frac{s}{m} \right)
	\\ & \textrm{with }m = \mathrm{SoftMin}_{e \in T}(\|e\|)
\end{align*}
The sign of the directed distances to the three edges indicates on which side of the edge a pixel is.
By applying a sigmoid function ($\sigma$) on that directed distances, we get a value close to 1 if the pixel lies inside and a value close to 0 if the pixel lies outside the triangle with respect to a given edge.
By taking the product of the values for all three edges, the result ($\in[0,1]$) smoothly indicates whether a pixel lies in or outside a triangle.
Since this draws the triangles only from one direction, we add the same term with the negative directed distances to make the visibility tensor triangle orientation invariant:
\begin{align*}
&V_\mathrm{orient.inv.}(p_x, p_y, T) =\\
&\quad \sum_{a\in \{-1,1\}} \prod_{e=(v_1, v_2)\in T}\sigma \left( a
\begin{vmatrix}
v_{x, 2}-v_{x, 1} & v_{x, 1}-p_{x} \\ 
v_{y, 2}-v_{y, 1} & v_{y, 1}-p_{y}
\end{vmatrix}
\frac{s}{m} \right) 
\end{align*}

For Figure \ref{fig:intriangle}, the visibility tensor $V$ looks as follows:
\begin{align*}
	V({\color{cyan}p_x}/{\color{magenta}p_x}, {\color{cyan}p_y}/{\color{magenta}p_y}, T) \overset{\text{example}}{=}\quad
	& \left(
	\raisebox{-1.35em}{\includegraphics[page=1, height=3em]{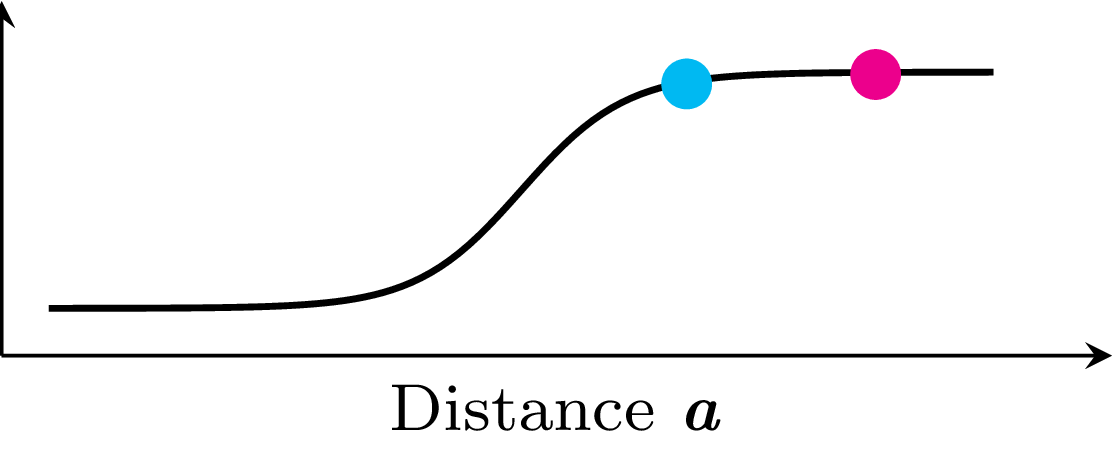}} \right)\\
	\cdot \left(
	\raisebox{-1.35em}{\includegraphics[page=2, height=3em]{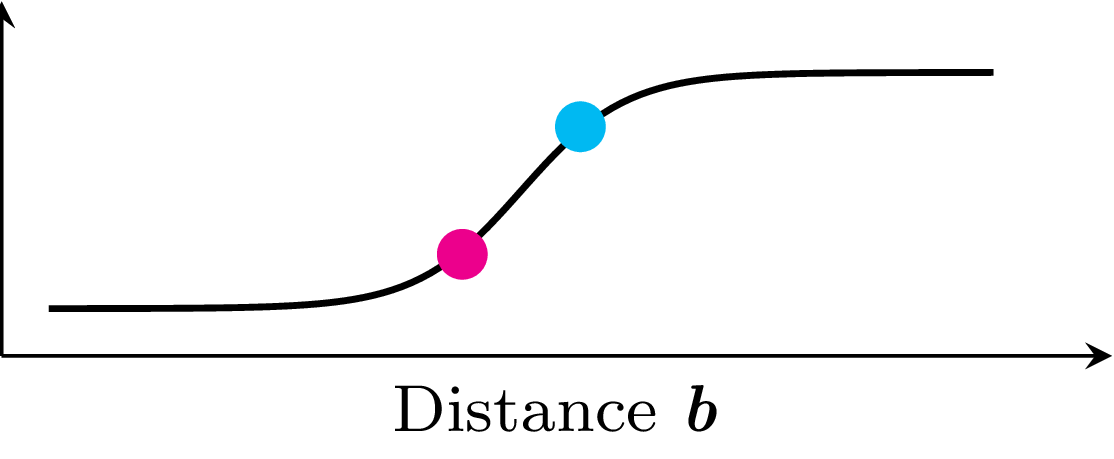}} \right)
	\cdot & \left(
	\raisebox{-1.35em}{\includegraphics[page=3, height=3em]{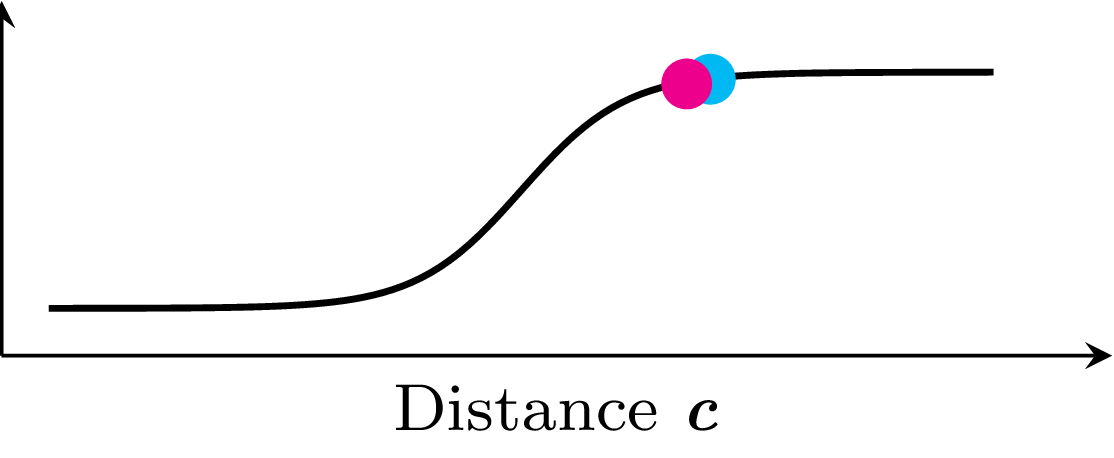}} \right)
\end{align*}

The steepness $s$ is a hyper-parameter setting the steepness of the sigmoid function. 
See Figure \ref{fig:renderings} to see how it affects the results.

\begin{figure*}
    \vspace{-5mm}
	\centering
	\begin{tabular}{ccc}
		\subcaptionbox{sub-RAN 1}{\includegraphics[width=.31\linewidth]{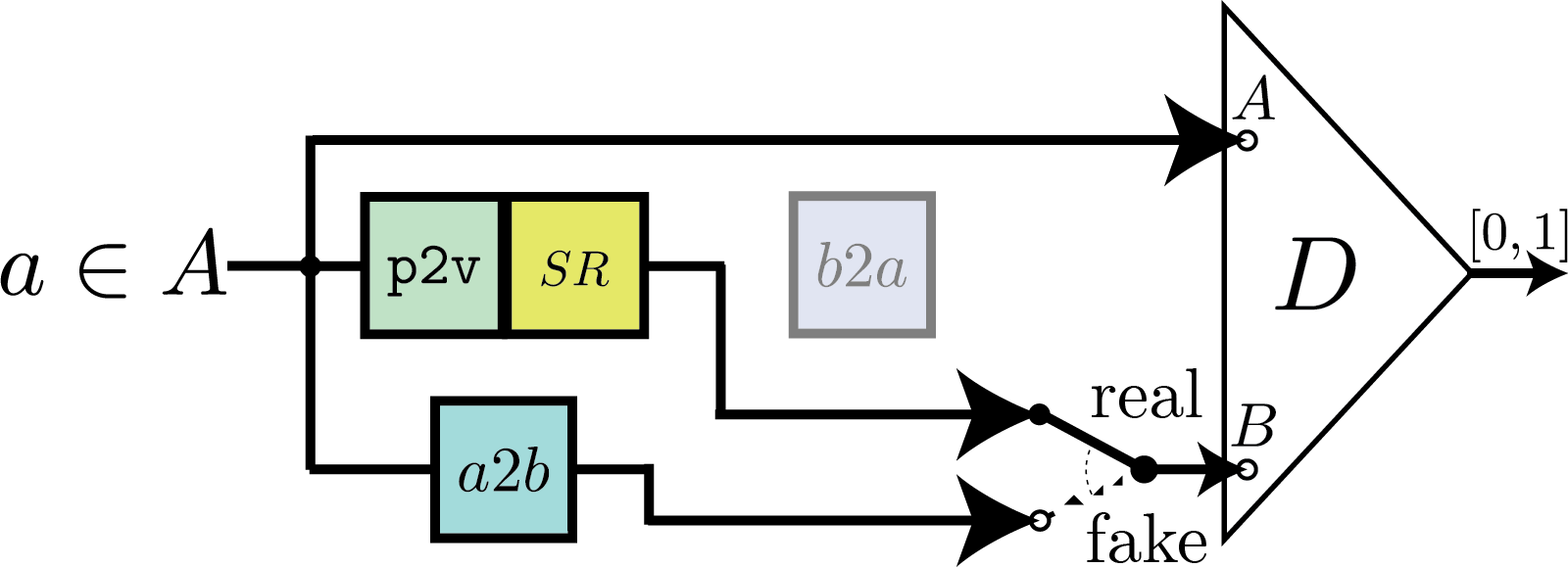}} &
		\subcaptionbox{sub-RAN 2}{\includegraphics[width=.31\linewidth]{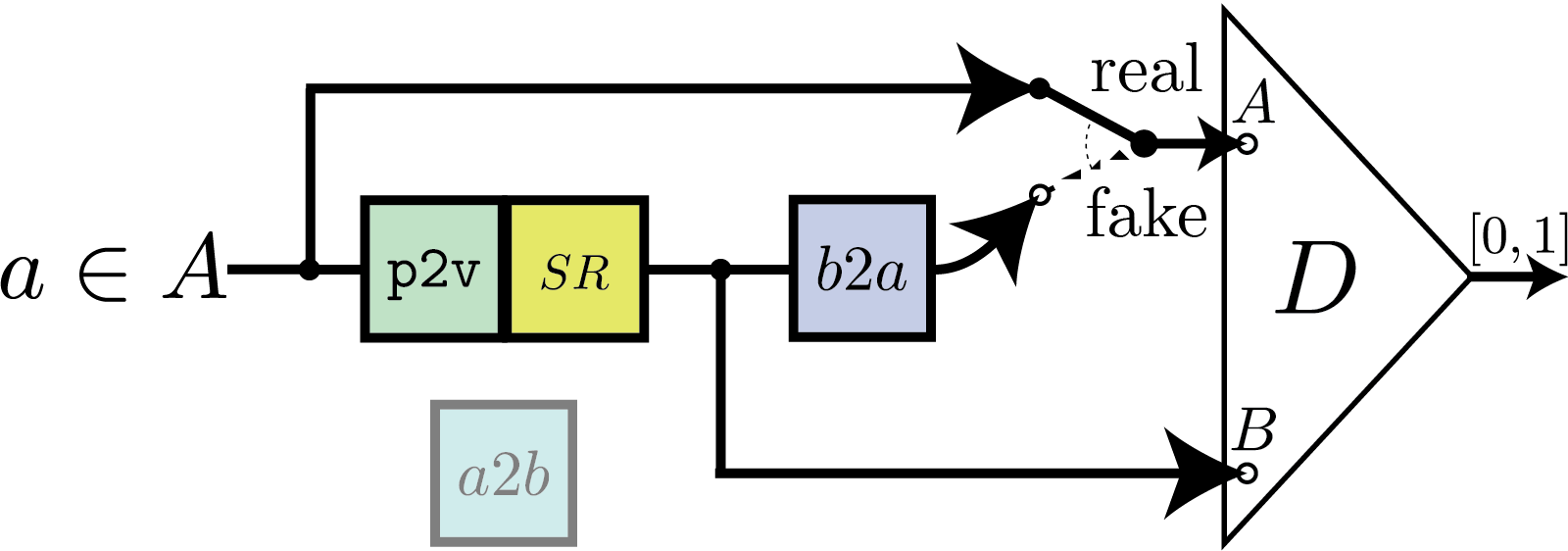}} &
		\subcaptionbox{sub-RAN 3}{\includegraphics[width=.31\linewidth]{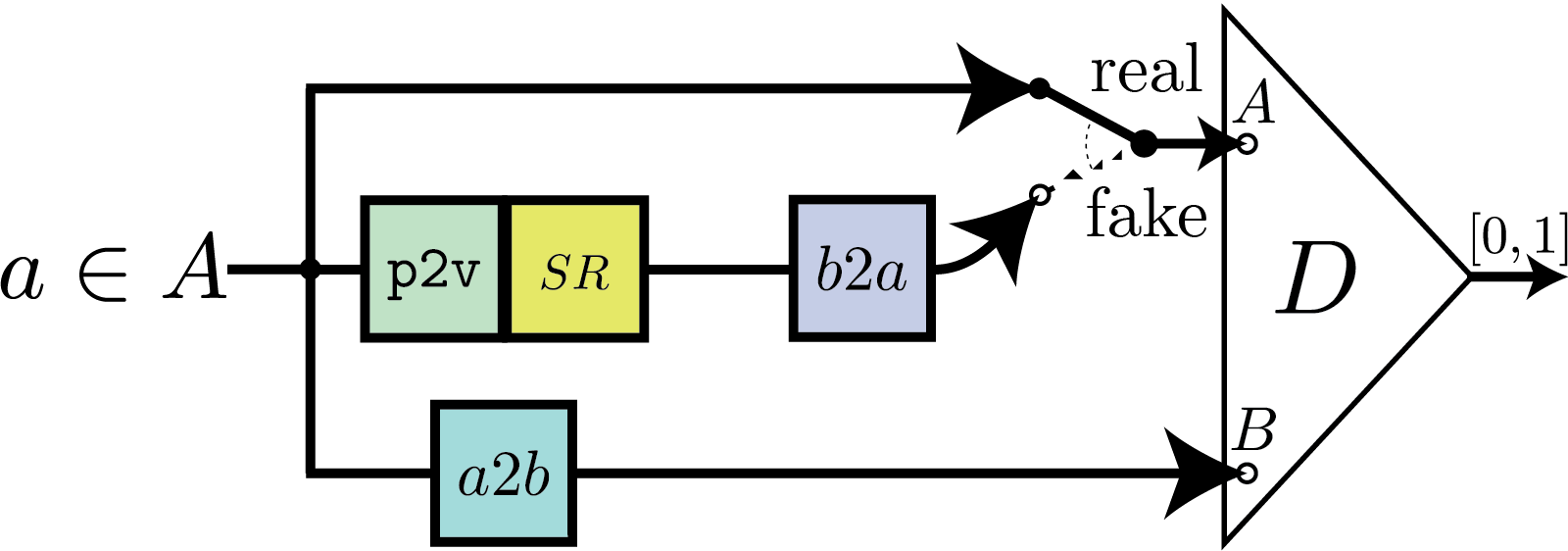}} \\
		~&
		\subcaptionbox{sub-RAN 4}{\includegraphics[width=.31\linewidth]{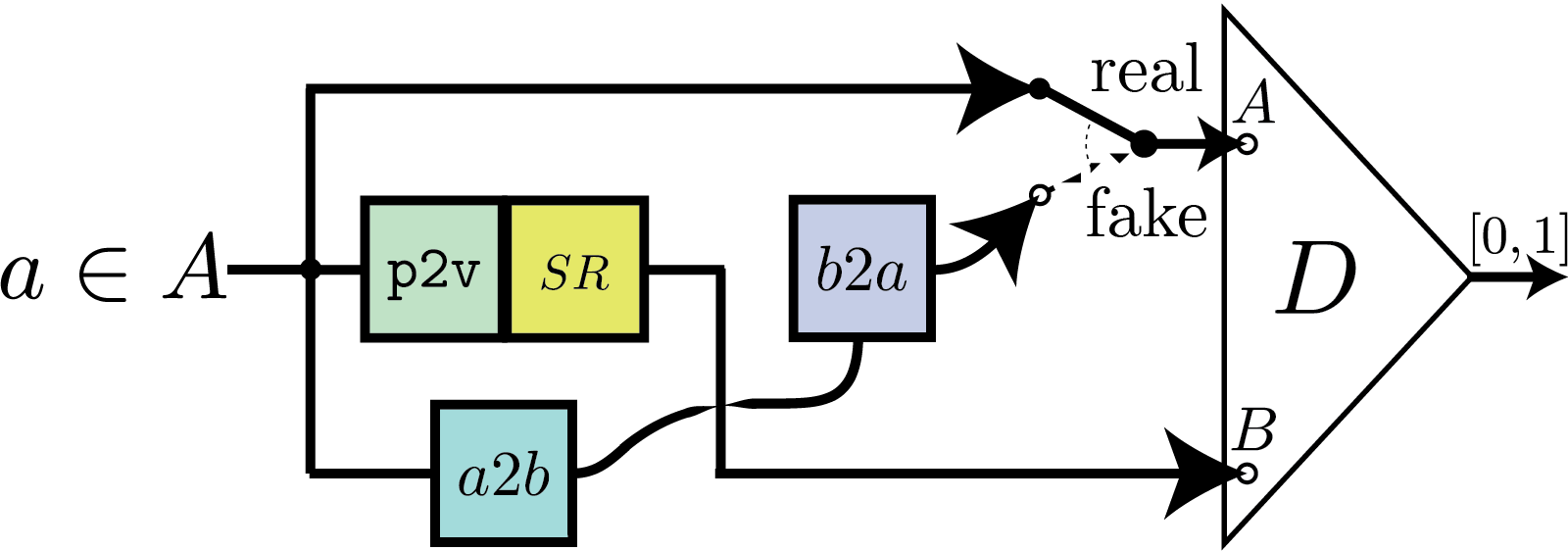}} &
		\subcaptionbox{sub-RAN 5}{\includegraphics[width=.31\linewidth]{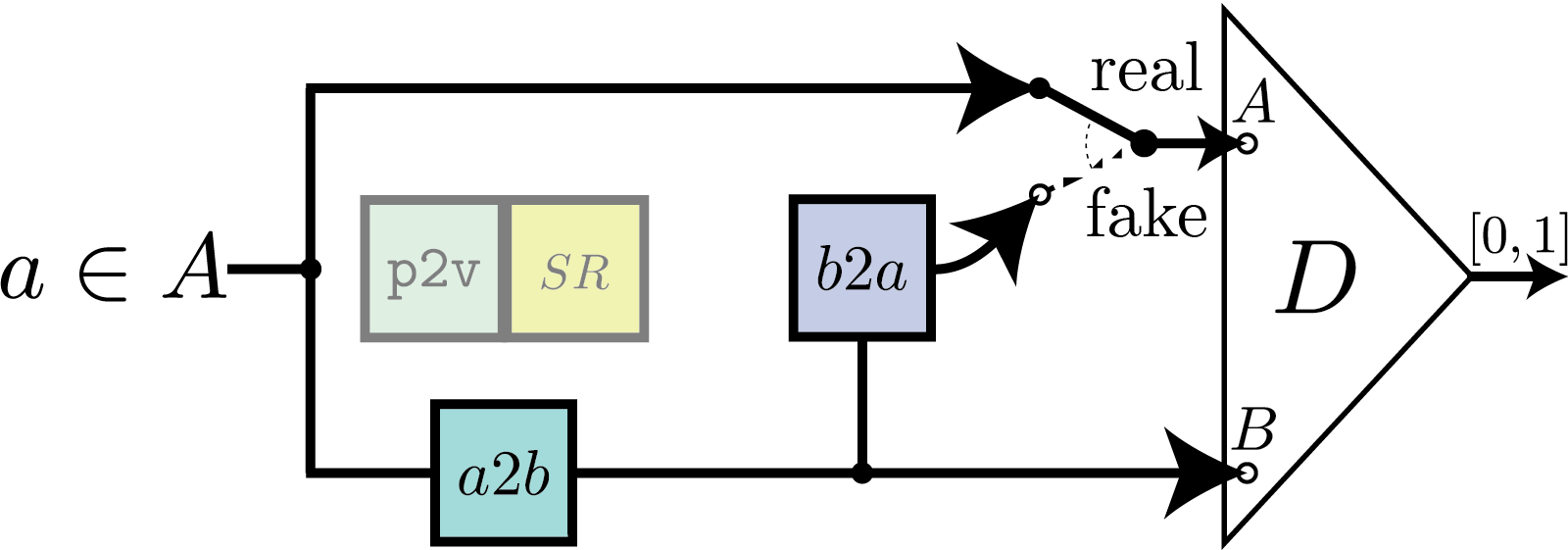}}
	\end{tabular}
	
	\justify
	
	\small\sffamily

	\vspace*{\dimexpr-\parskip-2.25cm\relax}%
	\parshape 8 %
	0pt 0.31\linewidth %
	0pt 0.31\linewidth %
	0pt 0.31\linewidth
	0pt 0.31\linewidth
	0pt 0.31\linewidth
	0pt 0.31\linewidth
	0pt 0.31\linewidth
	0pt \linewidth %
	\makeatletter
	\refstepcounter\@captype%
	\addcontentsline{\csname ext@\@captype\endcsname}{\@captype}%
	{\protect\numberline{\csname the\@captype\endcsname}{ToC entry}}%
	\csname fnum@\@captype\endcsname: %
	\makeatother
	The five sub-RANs constituting the RAN.
	Sub-RAN 1 (a) has a discriminator switch at input $B$, differentiating between \texttt{p2v}+$SR$ and $a2b$.
	The other four routes differentiate between two images from space $A$, while the ``real'' input for the discriminator is always the input image.
	The ``fake'' input is the result of a round trip that either involves \texttt{p2v}+$SR$ and $b2a$ (b, c) or $a2b$ and $b2a$ (d, e).
	This round trip is required for a cycle-consistency loss that diminishes mode collapse.
	Meanwhile, the $B$ input is the result of either \texttt{p2v}+$SR$ (b, d) or $a2b$ (c, e). 
	By that, \texttt{p2v} and $a2b$ can be trained mutually.
	\label{fig:5ran}
\end{figure*}

\begin{figure}
	\centering
	\hspace{2em}\includegraphics[width=.9\linewidth]{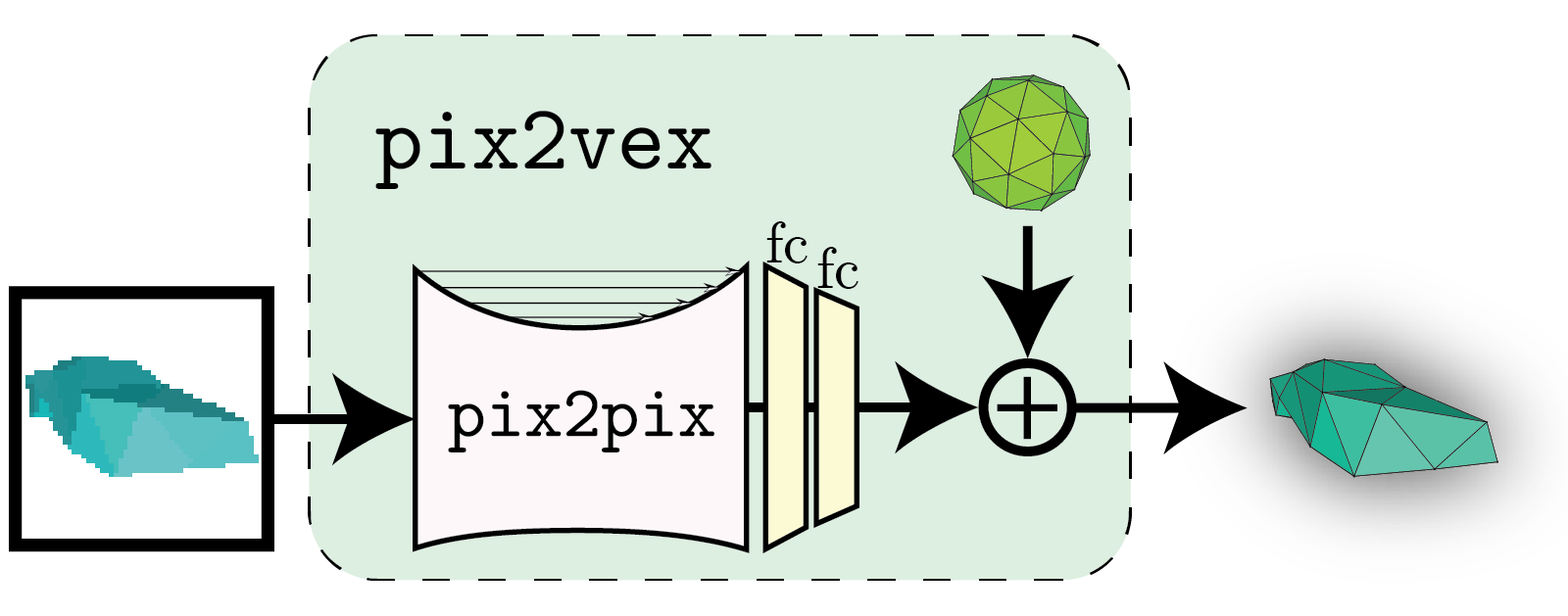}
	\caption{
		The reconstructor network's \texttt{pix2vex} architecture.
	}
	\label{fig:pix2vex}
\end{figure} %
\begin{figure}
	\centering
    \begin{tabular}{cc}
        \includegraphics[width=.45\linewidth]{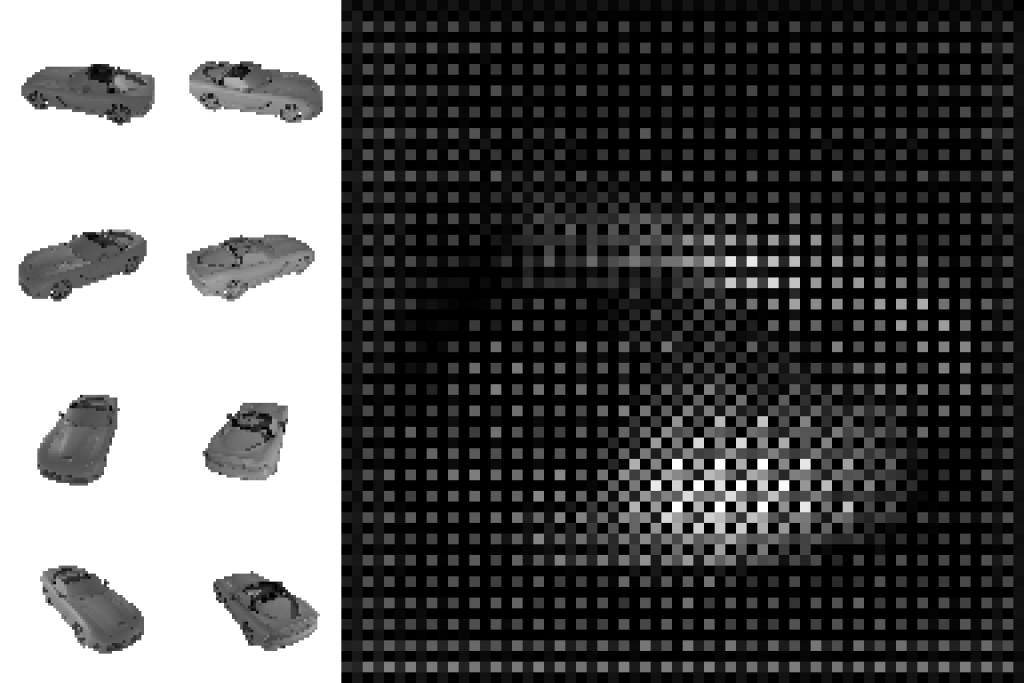} & 
        \includegraphics[width=.45\linewidth]{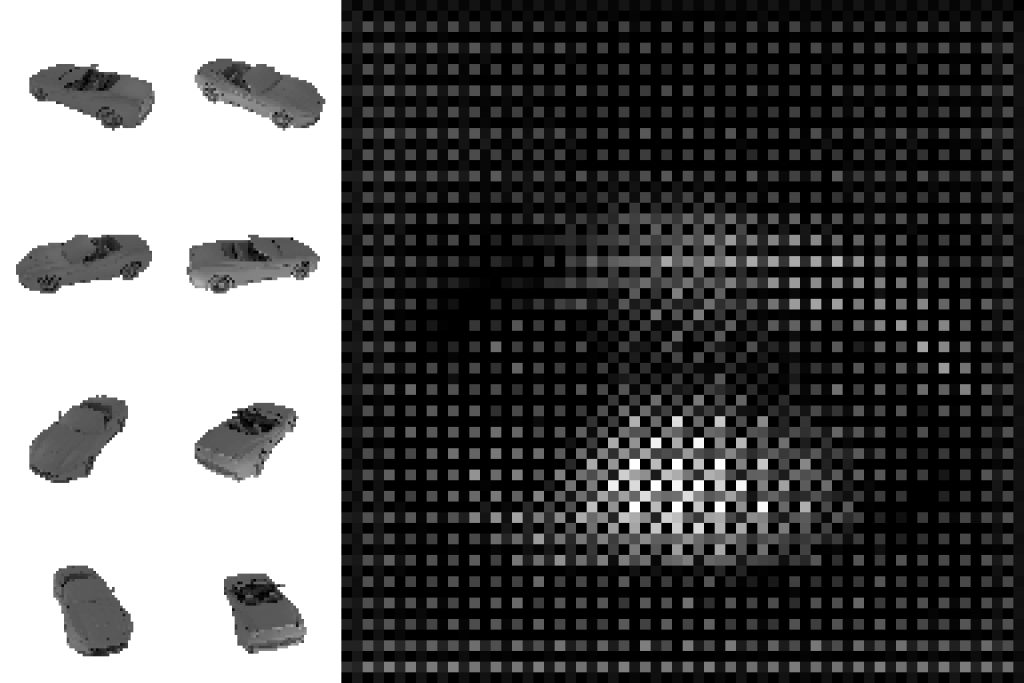} \\
        \includegraphics[width=.45\linewidth]{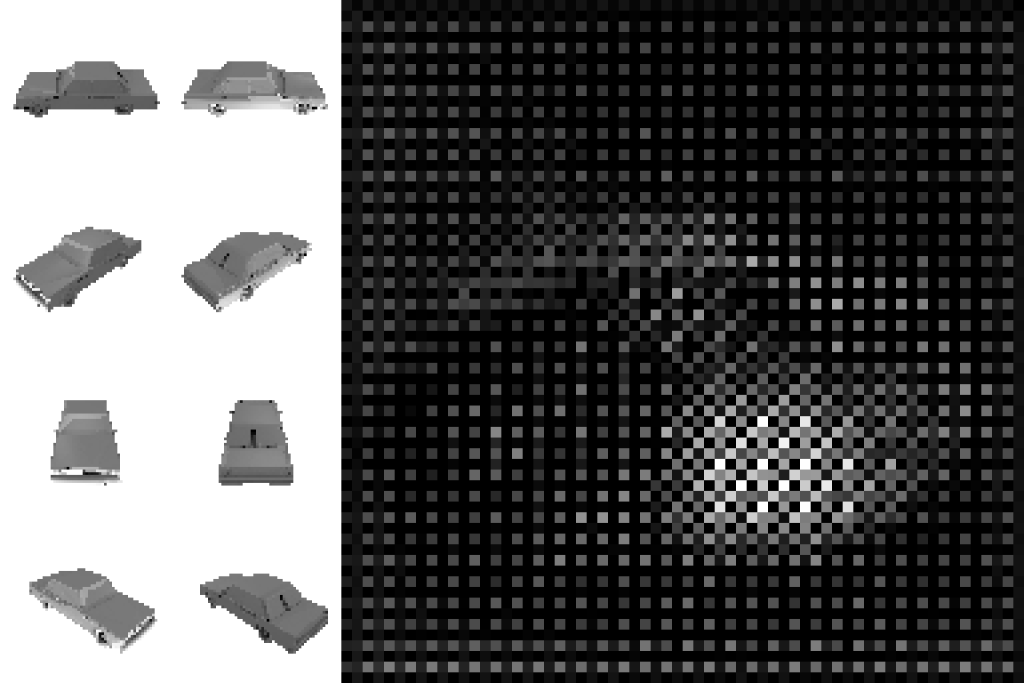} & 
        \includegraphics[width=.45\linewidth]{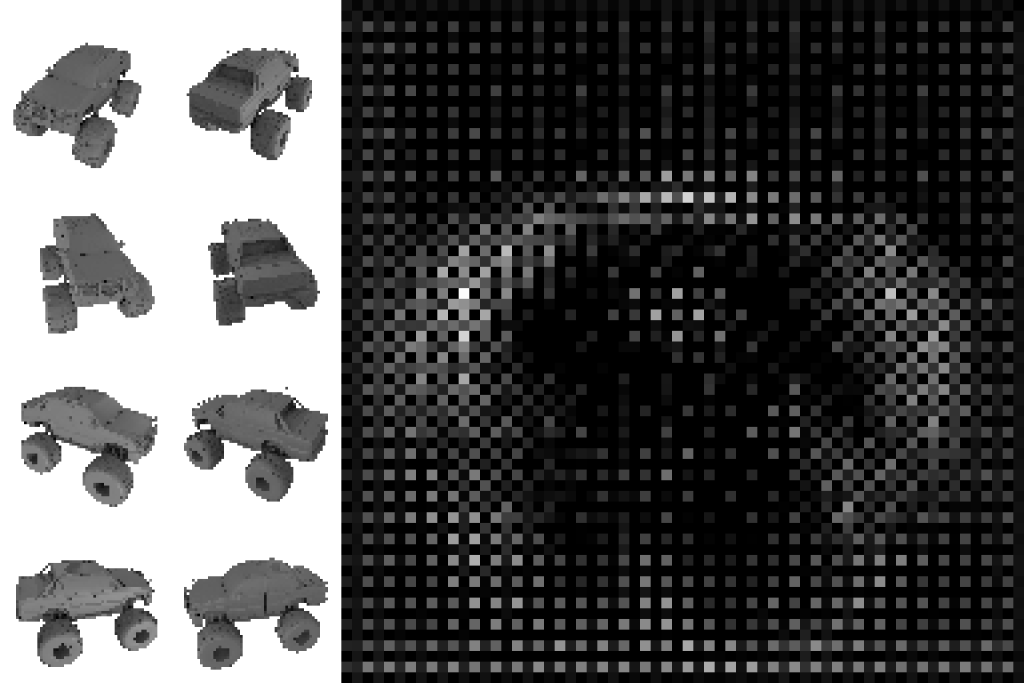} \\
        \includegraphics[width=.45\linewidth]{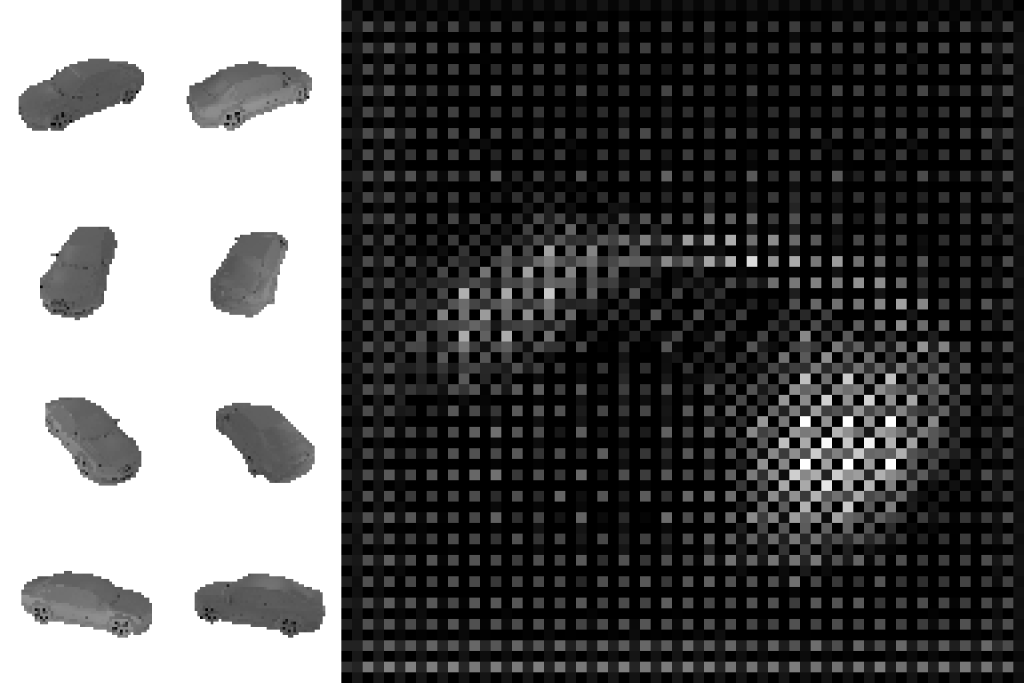} & 
        \includegraphics[width=.45\linewidth]{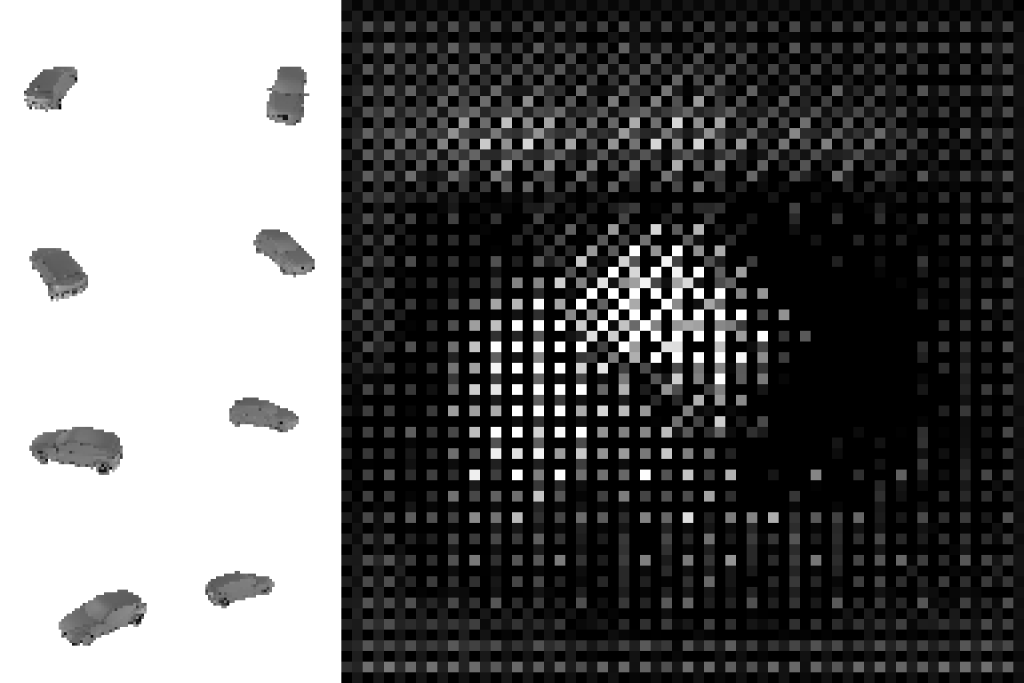}
    \end{tabular}
	\caption{Examples of the hidden states in the \texttt{pix2vex} architecture after the \texttt{pix2pix} component. In these cases, eight input images are transformed into a single representative image from which the 3D vertices can be deduced by two fully connected layers.}
	\label{fig:pix2vex_internal}
\end{figure} 
\section{Reconstructor (\texttt{pix2vex})}
\label{sub:reconstructor}\label{sec:pix2vex}

The reconstructor \texttt{pix2vex} (\texttt{p2v} in the figures) is the network that translates from images to a 3D model.
The view directions of the input images are for most of our experiments fixed.
The \texttt{pix2vex} architecture is based on a \texttt{pix2pix} convolutional and deconvolutional ResNet (\cite{He2015}) network.
The output of the \texttt{pix2pix} is followed by two fully connected layers bringing it to a size of $3\times \#\mathrm{vertices}$.
This is followed by a sigmoid layer and a linear transformation to bring the values into a range of $[-1, 1]$.
These values are considered as vectors and then added to the vertices of the base model, which is a uniform sphere with a predefined topology.
In our experiments, we have found that starting the training from an object, such as the sphere, and predicting offsets from it, evolves in a better and smoother way towards convergence in comparison to predicting absolute coordinates.%

To reconstruct models that contain symmetry we can optionally use a more canonical way of displacing vertices from an initial model to the prediction.
That is, we split our model into four equal quarters w.r.t.{} a top-down view.
Firstly, we predict a quarter of the model by applying shifts to the base model of that quarter.
Then we mirror the predicted quarter to another quarter and predict the shifts to displace the mirrored quarter.
We do this until all four quarters are predicted.
This implies the gradients that would usually only move one side to imply a movement on all sides.

In addition, we also predict uniform translation and scaling values, which assist in reducing the overall offset values magnitude, and thus improve system stability.

Lastly, we also apply regularization on the 3D mesh, which guide the prediction towards more reasonably shaped reconstructed meshes.
These regularizations consider angles between normals, edge lengths, and curvature, as shall be described in more detail in Appendix \ref{apx:losses}.

Compared to other networks that predict vertices from one or more images, our network has an image as its native internal state. Thus, it allows for a stronger interpretability of the inner working of our neural network. Figure.~\ref{fig:pix2vex_internal} presents the internal states corresponding to the respective inputs.

\begin{figure*}[t]
    \renewcommand{\imageone}[1]{\hspace{-.9em}\raisebox{.2cm}{\includegraphics[width=.098\linewidth, ]{images/classic/#1.png}}\hspace{-.9em}}
    \renewcommand{\imagetwo}[1]{\hspace{-.9em}\raisebox{.0cm}{\includegraphics[width=.098\linewidth, ]{images/classic/#1.png}}\hspace{-.9em}}
	\centering
	{\scriptsize
	\begin{tabular}{cccccccccc}
	    \hspace{-1em}Input\hspace{-2em} & & & & \hspace{-1em}Prediction\hspace{-1em} & & & & \hspace{-1em}Alternative~View\hspace{-2em} & \\
	    \imageone{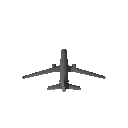} &
	    \imageone{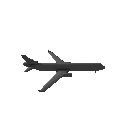} &
	    \imageone{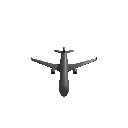} &
	    \imageone{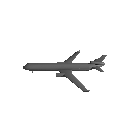} &
	    \imagetwo{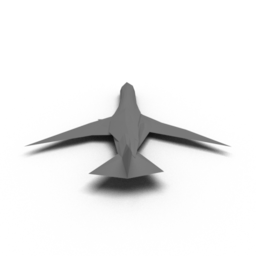} &
	    \imagetwo{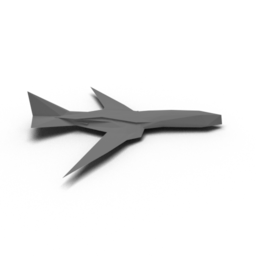} &
	    \imagetwo{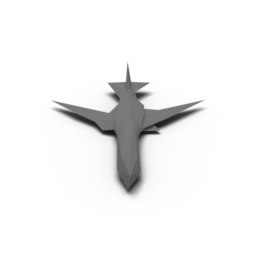} &
	    \imagetwo{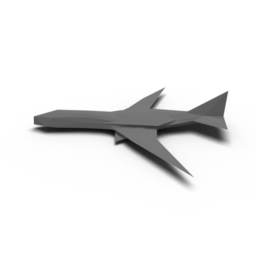} &
	    \imagetwo{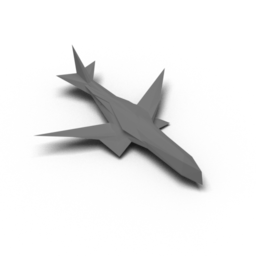} &
	    \imagetwo{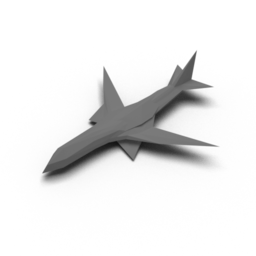} \\[-.2cm] %
	    \imageone{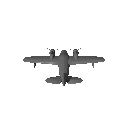} &
	    \imageone{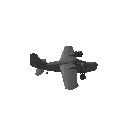} &
	    \imageone{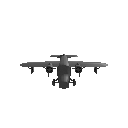} &
	    \imageone{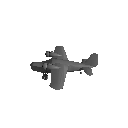} &
	    \imagetwo{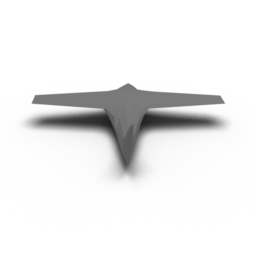} &
	    \imagetwo{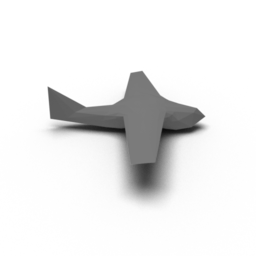} &
	    \imagetwo{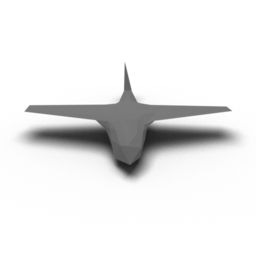} &
	    \imagetwo{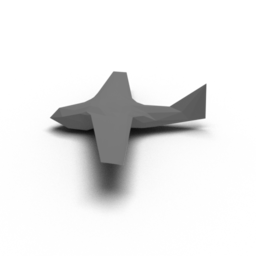} &
	    \imagetwo{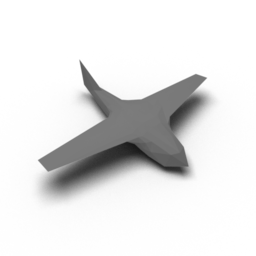} &
	    \imagetwo{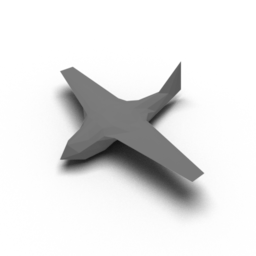} \\[-.2cm] %
	    \imageone{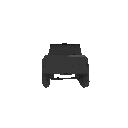} &
	    \imageone{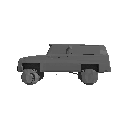} &
	    \imageone{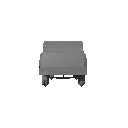} &
	    \imageone{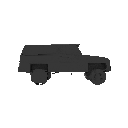} &
	    \imagetwo{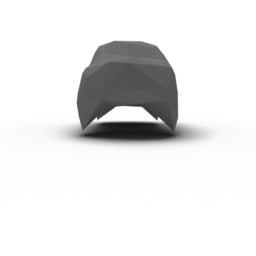} &
	    \imagetwo{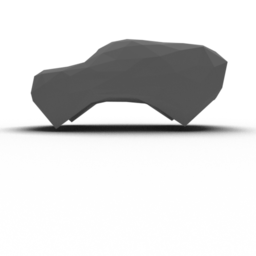} &
	    \imagetwo{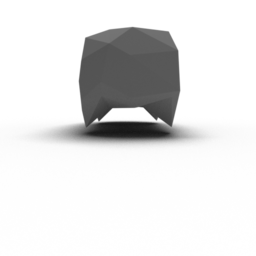} &
	    \imagetwo{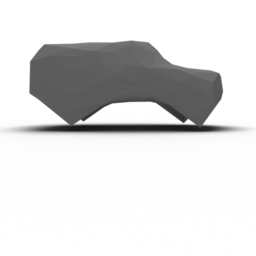} &
	    \imagetwo{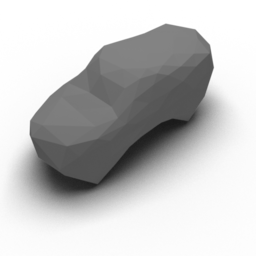} &
	    \imagetwo{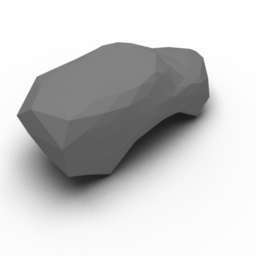} \\[-.2cm] %
	    \imageone{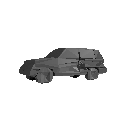} &
	    \imageone{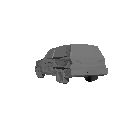} &
	    \imageone{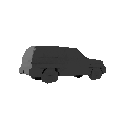} &
	    \imageone{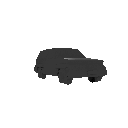} &
	    \imagetwo{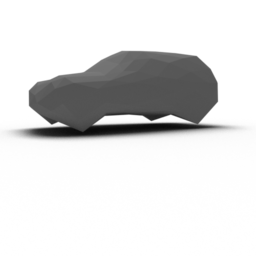} &
	    \imagetwo{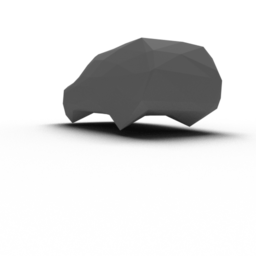} &
	    \imagetwo{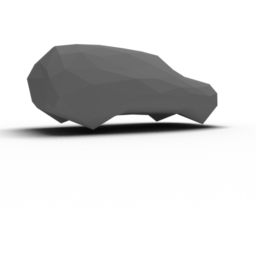} &
	    \imagetwo{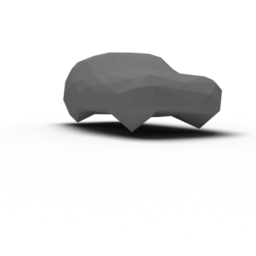} &
	    \imagetwo{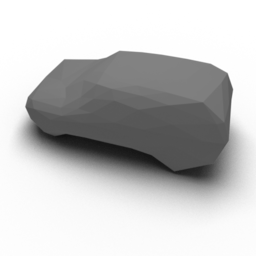} &
	    \imagetwo{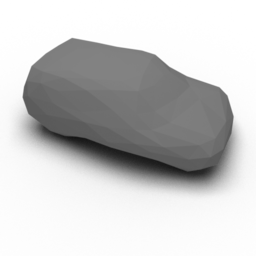} \\[-.2cm] %
	    \imageone{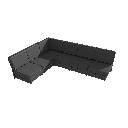} &
	    \imageone{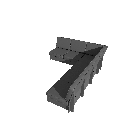} &
	    \imageone{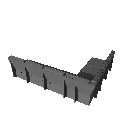} &
	    \imageone{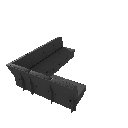} &
	    \imagetwo{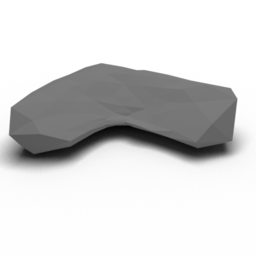} &
	    \imagetwo{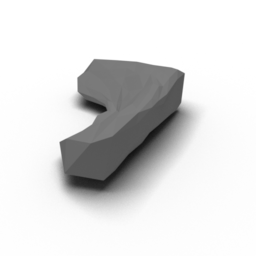} &
	    \imagetwo{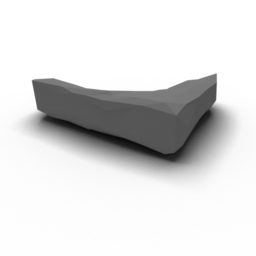} &
	    \imagetwo{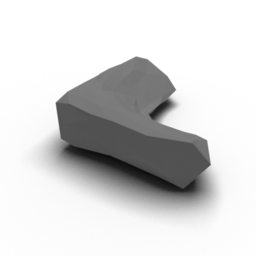} &
	    \imagetwo{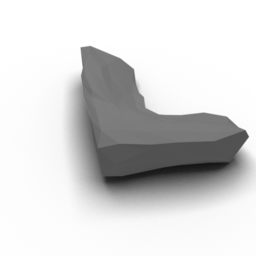} &
	    \imagetwo{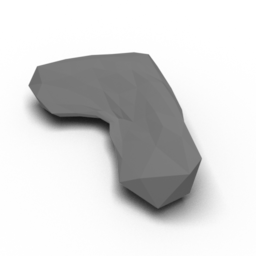} \\[-.2cm] \\ %
	\end{tabular}}
	{\scriptsize
	\begin{tabular}{cccccccccc}
	    \hspace{-1em}Input (only 2 of 4)\hspace{-5em} & & \hspace{-1em}Prediction (only 2 of 4)\hspace{-7em} & & \hspace{-1em}Alternative View\hspace{-1em} &
	    \hspace{-1em}Input (only 2 of 4)\hspace{-5em} & & \hspace{-1em}Prediction (only 2 of 4)\hspace{-7em} & & \hspace{-1em}Alternative View\hspace{-1em} \\
	    \imageone{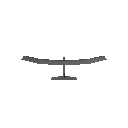} &
	    \imageone{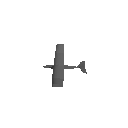} &
	    \imagetwo{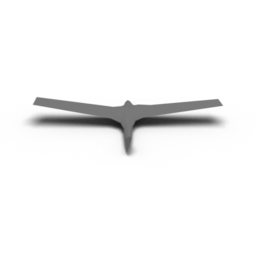} &
	    \imagetwo{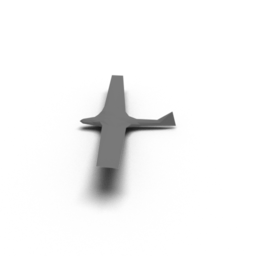} &
	    \imagetwo{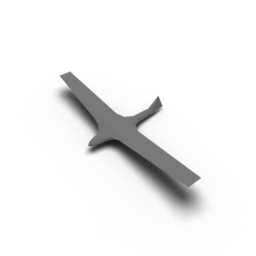} &
	    \imageone{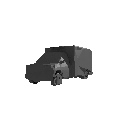} &
	    \imageone{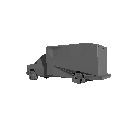} &
	    \imagetwo{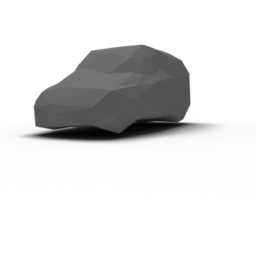} &
	    \imagetwo{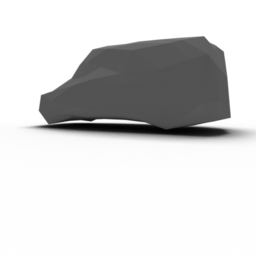} &
	    \imagetwo{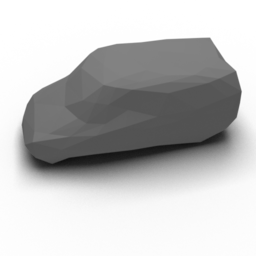} \\[-.2cm] %
	    \imageone{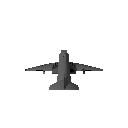} &
	    \imageone{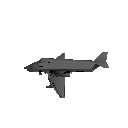} &
	    \imagetwo{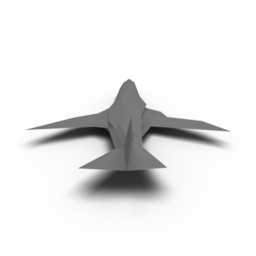} &
	    \imagetwo{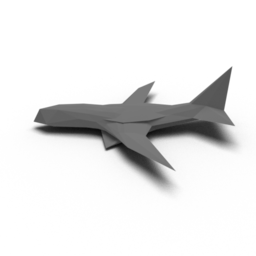} &
	    \imagetwo{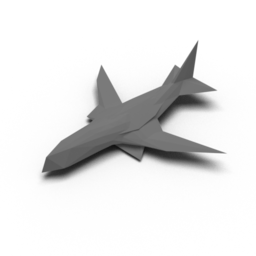} &
	    \imageone{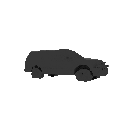} &
	    \imageone{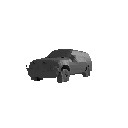} &
	    \imagetwo{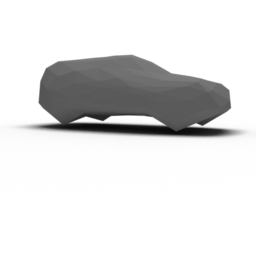} &
	    \imagetwo{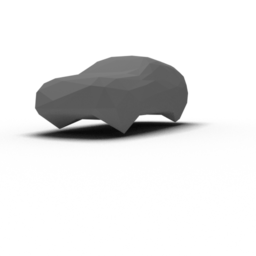} &
	    \imagetwo{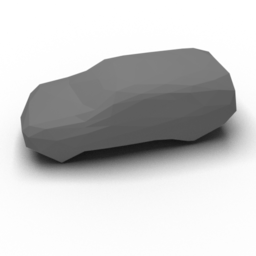} \\[-.2cm] %
	\end{tabular}}
	\captionsetup{width=.95\linewidth}
	\caption{
	    Four-view reconstruction of the ShapeNet \cite{Chang2015} classes \textit{airplane}, \textit{car}, and \textit{sofa}.
	    {Left:} input. {Middle:} prediction from the same angles. {Right:} predictions from alternative viewpoints.
	\label{fig:results-classic}}
\end{figure*}

\begin{figure}[htb]
    \renewcommand{\imageone}[1]{\makecell[t]{\hspace{-.75em}\includegraphics[width=.16\linewidth, ]{images/shoes/#1}\hspace{-.75em}}}
    \renewcommand{\imagetwo}[1]{\makecell{\hspace{-.75em}\hspace{-.08\linewidth}\includegraphics[width=.29\linewidth, ]{images/shoes/#1}\hspace{-.08\linewidth}\hspace{-.75em}}}
	\centering
	\setcellgapes{3pt}\makegapedcells
	\vspace{-.05cm}
	{\scriptsize
	\begin{tabular}{ccccccc}
	    Input & Pred. & Alt.~v. &~& Input & Pred. & Alt.~v. \\
	    \imageone{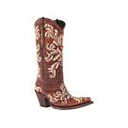} &
	    \imagetwo{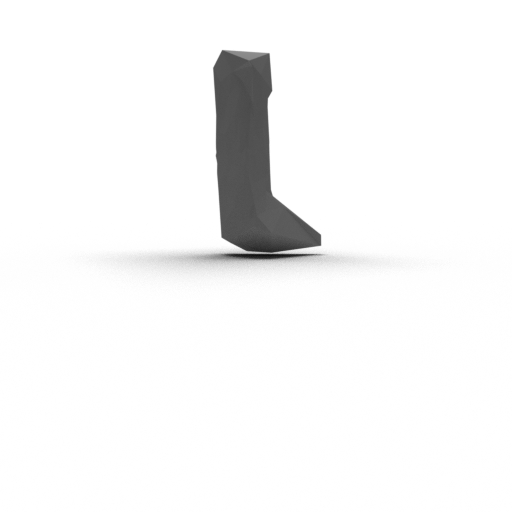} &
	    \imagetwo{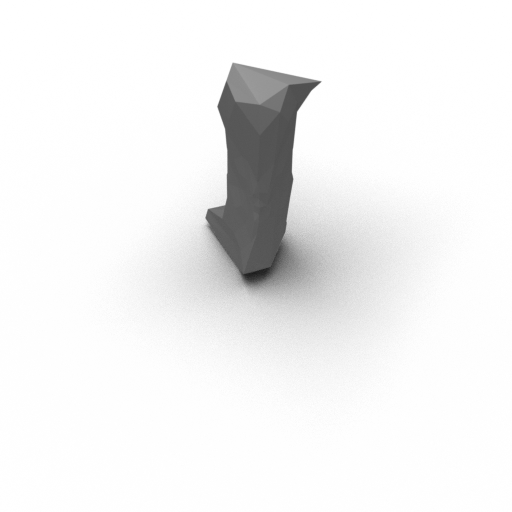} &&
	    \imageone{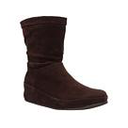} &
	    \imagetwo{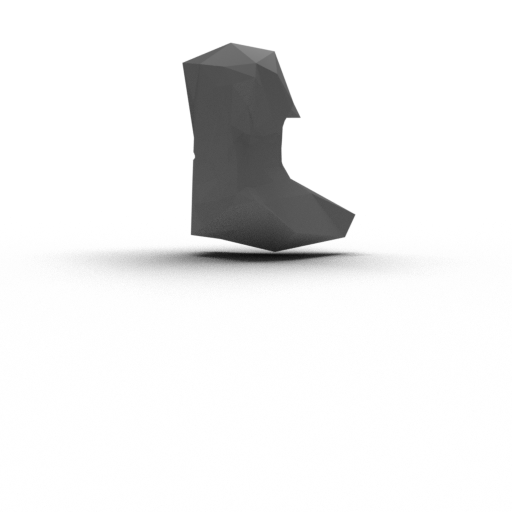} &
	    \imagetwo{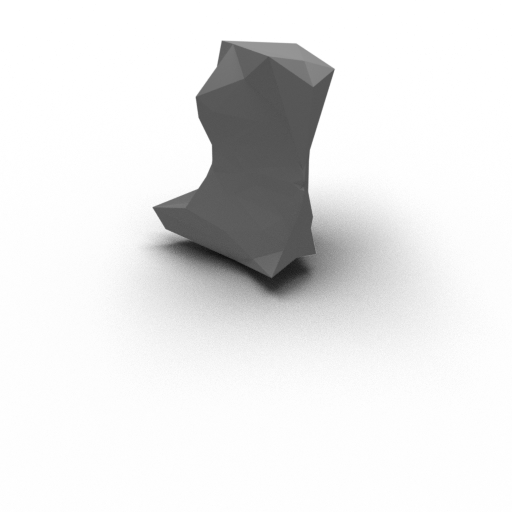} \\[-.13\linewidth] %
	    \imageone{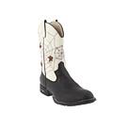} &
	    \imagetwo{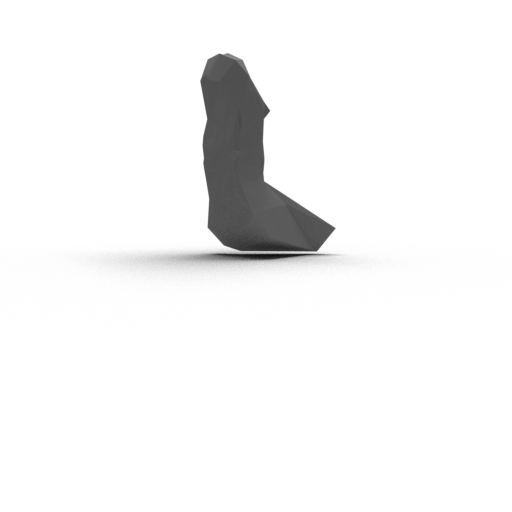} &
	    \imagetwo{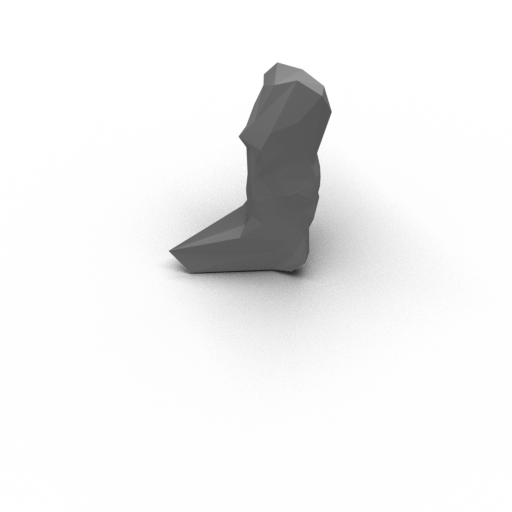} &&
	    \imageone{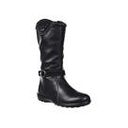} &
	    \imagetwo{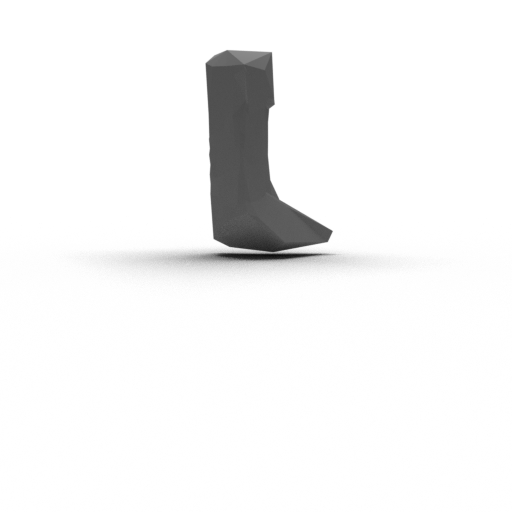} &
	    \imagetwo{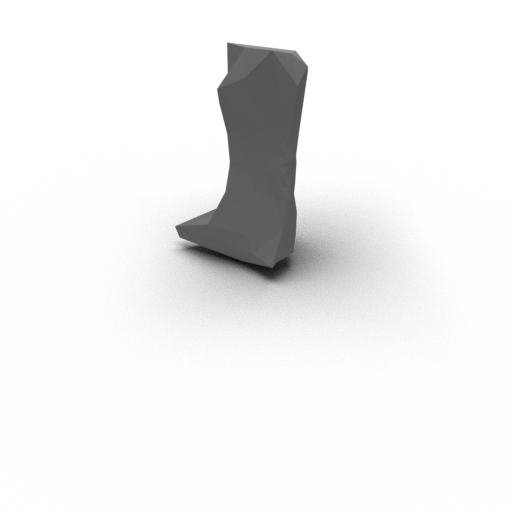} \\[-.13\linewidth] %
	    \imageone{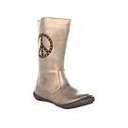} &
	    \imagetwo{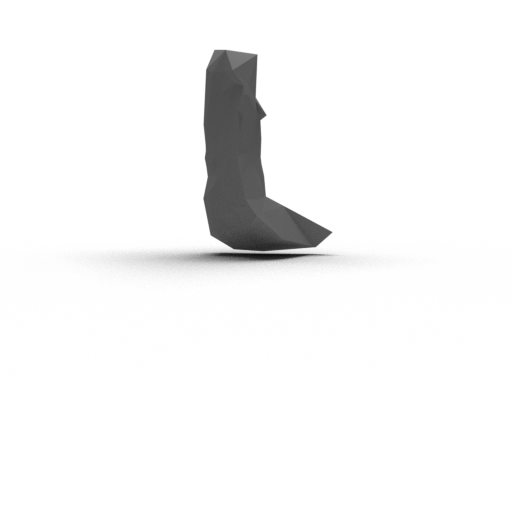} &
	    \imagetwo{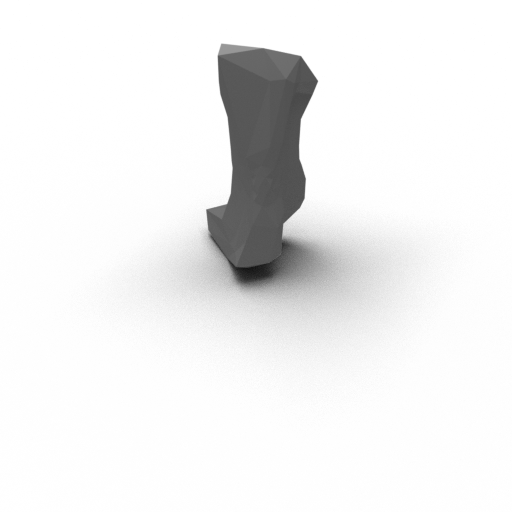} &&
	    \imageone{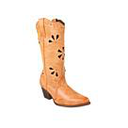} &
	    \imagetwo{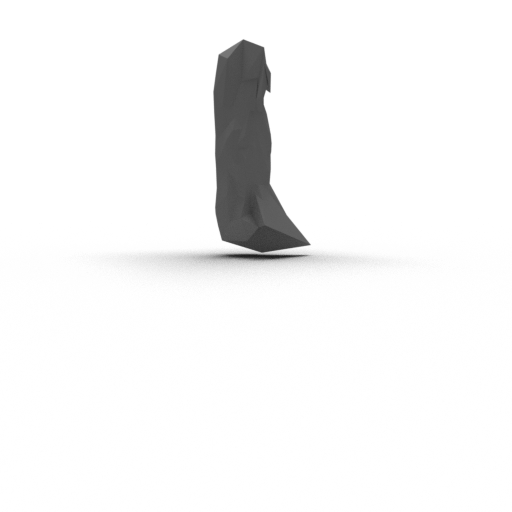} &
	    \imagetwo{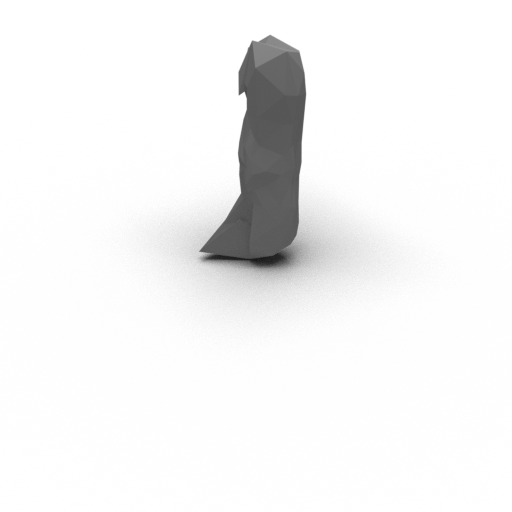} \\[-.16\linewidth]\\ %
	\end{tabular}}
	\caption{
	Single-view reconstruction results from the UT Zappos50K dataset (camera-captured images).
	\label{fig:results-shoes}}
\end{figure}

\begin{figure}[htb]
    \renewcommand{\imageone}[1]{\hspace{-1.1em}\makecell{\raisebox{.3cm}{\includegraphics[width=.175\linewidth, ]{images/exp/#1}}\hspace{-1.1em}}}
    \renewcommand{\imagetwo}[1]{\hspace{-1.1em}\makecell{\raisebox{.0cm}{\includegraphics[width=.16\linewidth, ]{images/exp/#1}}\hspace{-1.1em}}}
	\centering
	{\scriptsize
	\begin{tabular}{cccccc}
	    & \hspace{-1em}Input (only 2 of 4)\hspace{-5em} & & \hspace{-1em}Pred. (only 2 of 4)\hspace{-4em} & & \hspace{-1em}Alt.~v.\hspace{-1em} \\ %
	    \makecell{(a)} &
	    \imageone{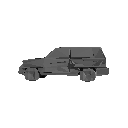} &
	    \imageone{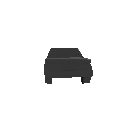} &
	    \imagetwo{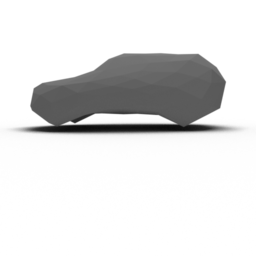} &
	    \imagetwo{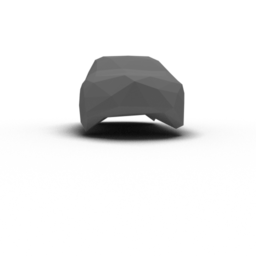} &
	    \imagetwo{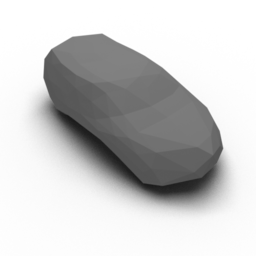} \\[-.25cm] %
	    \makecell{(b)} &
	    \imageone{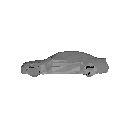} &
	    \imageone{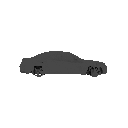} &
	    \imagetwo{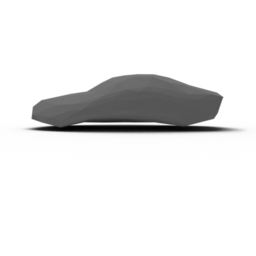} &
	    \imagetwo{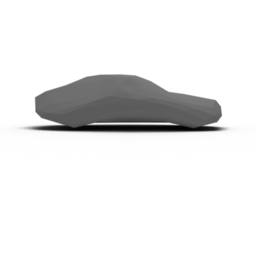} &
	    \imagetwo{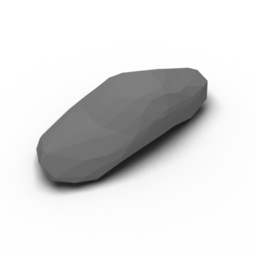} \\[-.25cm] %
	    \makecell{(c1)} &
	    \imageone{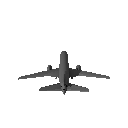} &
	    \imageone{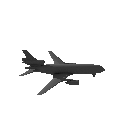} &
	    \imagetwo{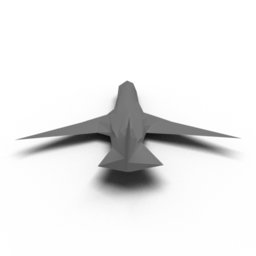} &
	    \imagetwo{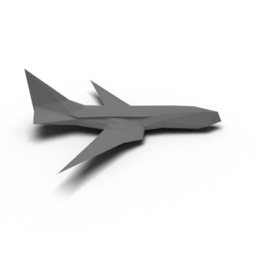} &
	    \imagetwo{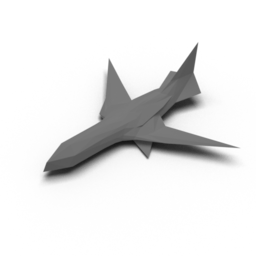} \\[-.25cm] %
	    \makecell{(c2)} &
	    \imageone{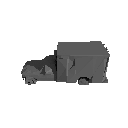} &
	    \imageone{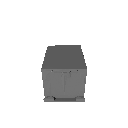} &
	    \imagetwo{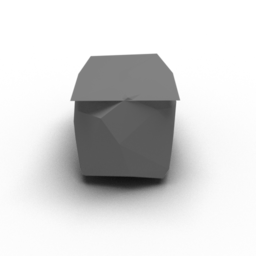} &
	    \imagetwo{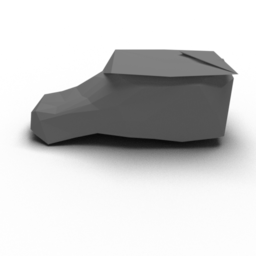} &
	    \imagetwo{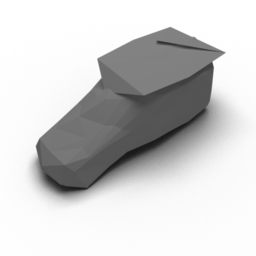} \\
	    & \hspace{-1em}Input\hspace{-1em} & & \hspace{-1em}Pred.\hspace{-1em} & & \hspace{-1em}Alt.~v.\hspace{-1em} \\ %
	    \makecell{(d1)} &
	    \imageone{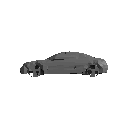} &
	    \imageone{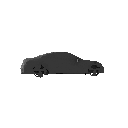} &
	    \imagetwo{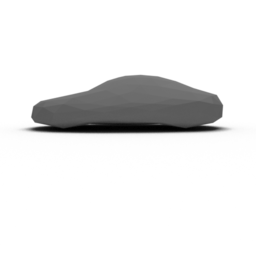} &
	    \imagetwo{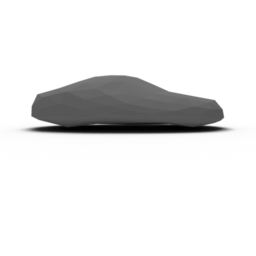} &
	    \imagetwo{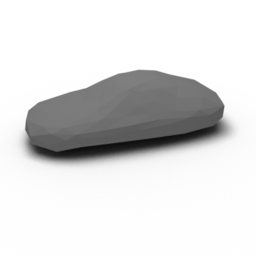} \\[-.25cm] %
	    \makecell{(d2)} &
	    \imageone{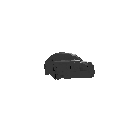} &
	    \imageone{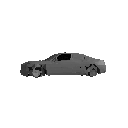} &
	    \imagetwo{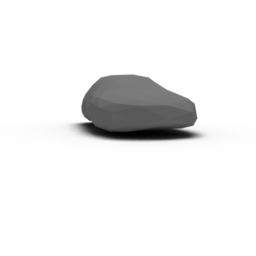} &
	    \imagetwo{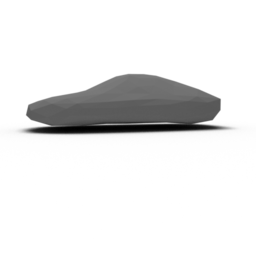} &
	    \imagetwo{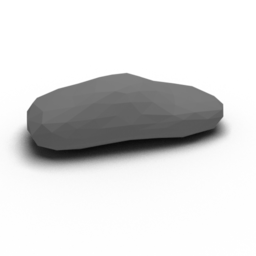} \\
	\end{tabular}}
    \renewcommand{\imageone}[1]{\hspace{-.9em}\makecell{\raisebox{.3cm}{\includegraphics[width=.147\linewidth, ]{images/exp/#1}}}\hspace{-.9em}}
    \renewcommand{\imagetwo}[1]{\hspace{-.9em}\makecell{\includegraphics[width=.147\linewidth, ]{images/exp/#1}}\hspace{-.9em}}
	{\scriptsize
	\begin{tabular}{cccccccc}
	    & \hspace{-1em}Input\hspace{-1em} & \hspace{-1em}Pred.\hspace{-1em} & \hspace{-1em}Alt.~v.\hspace{-1em} &
	    ~&
	    \hspace{-1em}Input\hspace{-1em} & \hspace{-1em}Pred.\hspace{-1em} & \hspace{-1em}Alt.~v.\hspace{-1em} \\ %
	    \makecell{(e1)\\(e2)} &
	    \imageone{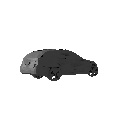} &
	    \imagetwo{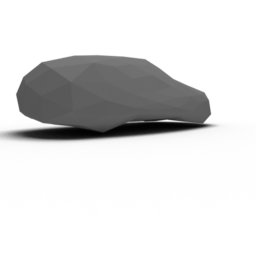} &
	    \imagetwo{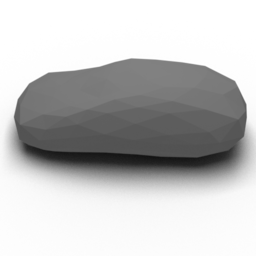} &&
	    \imageone{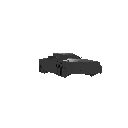} &
	    \imagetwo{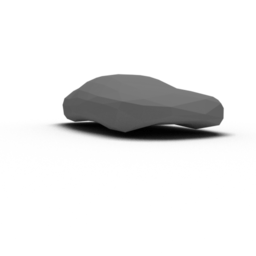} &
	    \imagetwo{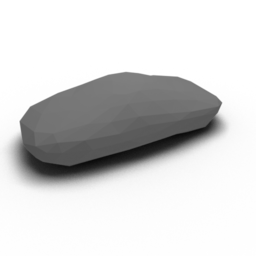} \\[-.25cm] %
	    \makecell{(e3)\\(e4)} &
	    \imageone{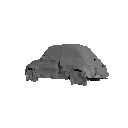} &
	    \imagetwo{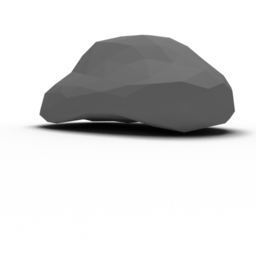} &
	    \imagetwo{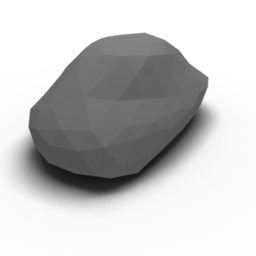} &&
	    \imageone{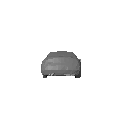} &
	    \imagetwo{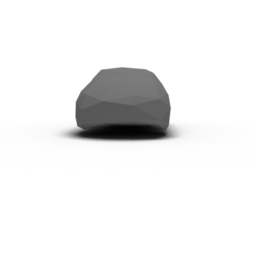} &
	    \imagetwo{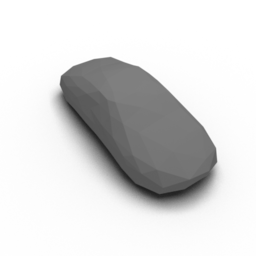} \\[-.25cm] %
	\end{tabular}}
	\caption{
	    Experiments showing the robustness of our approach.
	    (a--c): four-view training with the following modifications to the training data:
	    (a) randomized azimuth of the images; (b) randomly assigned position of the light source; (c) simultaneous training of \textit{car} and \textit{airplane} classes; (d) predicting images from only two input images; (e) single-view reconstructions.}
	\label{fig:results-exp}
\end{figure} 
\section{RAN}\label{sec:ran}

Our Reconstructive Adversarial Network (RAN) is a framework to train the reconstructor (\texttt{pix2vex}) without 3D supervision, i.e., without the need for the actual 3D models corresponding to the input images during training.
This means that we need another way to evaluate the predictions.
In other words, we let the RAN learn a supervision by itself.
The challenge here is the following causality dilemma:
To train the reconstruction, we need to compare the smoothly rendered images of the predicted shape to the input, which requires style transfer.
On the other hand, in order to train the style translation, we need to know what a properly smoothly rendered image (corresponding to the input image) looks like.
A typical approach for solving such causality dilemmas is to solve the two components coevolutionarily by iteratively applying various influences towards a common solution.

The key idea is to train an adversarial discriminator $D$ to discriminate between the different ways to obtain pairs of images from $A$ (identity, \texttt{p2v}--$SR$--$b2a$, $a2b$--$b2a$) and $B$ (\texttt{p2v}--$SR$, $a2b$).
This allows the three components \texttt{p2v}, $a2b$ and $b2a$ to be trained to fool $D$.
In designing such a strategy, we exploited the following insights:
\begin{itemize}[leftmargin=*]
    \item Since \texttt{pix2pix} networks are lazy and their capabilities are restricted, the discriminator can be implicitly trained in a way that the content between pairs of images ($A$ and $B$) will be similar.
        The rationale behind this is the following: for the \texttt{pix2pix} to hold the cycle-consistency of \texttt{p2v}--$SR$--$b2a$, it is much easier for the image translator to only do a style-transfer from a content-wise similar image than to reconstruct the input from a smooth rendering of a different object.

    \item To let the discriminator know what a general smoothly rendered image should look alike, we train it by rendering randomly guessed 3D models.
		After doing so, the discriminator can be used to train $a2b$ to output images from $B$.
\end{itemize}

Training a conventional GAN is a relatively straightforward task, since only a single binary decision (real vs.{} fake) has to be taken. 
Training the RAN, as shown in Fig. \ref{fig:rancomb}, is much more convoluted, since instead of only a single binary decision two decisions have to be made: 
one between three choices ($A$-input real vs.~fake generated by \texttt{p2v}--$SR$--$b2a$ vs.{} $a2b$--$b2a$) and one between two choices ($B$-input only fake generated by \texttt{p2v}--$SR$ vs.{} $a2b$).
These paths represent all possibilities to obtain an image of $A$ respectively $B$---i.e., the discriminator has to differentiate between all possible ways to generate its input and thus fooling the discriminator leads to a common solution for all these paths.

Since it is easier to train a binary discriminator, our solution is to break the RAN into five sub-RANs,
which all have to take only a single binary decision (real vs.~fake), as depicted in Figure \ref{fig:5ran}.
These sub-RANs are alternately trained like conventional GANs by training their discriminator to differentiate between the ``real'' and ``fake'' input and training the other modules to fool the discriminator.
If there are two modules to be trained at once, the training is split into two steps: 
the module next to the discriminator is trained first and the one after the input is trained second
(e.g., $b2a$, which is close to the discriminator, is trained first, and $a2b$ is trained second).
This helps to avoid mode collapse.
Since the relevance of these sub-RANs differs, their influence is weighted.
For example, training a path with the \texttt{pix2vex} module (sub-RAN (a) in Figure \ref{fig:5ran}) carries more weight than training the cycle of the two image translators (sub-RAN (e)).
For the discriminator $D$, we use the binary cross-entropy loss.
For training, an $L^1$ loss between any two images of the same image space is applied (details in Appendix \ref{apx:losses}).

This constitutes the RAN as an unsupervised way to find an appropriate internal representation which in turn requires the \texttt{pix2pix} networks to perform a minimum of content-wise changes.

\section{Results}\label{sec:res}

We evaluate our reconstruction results on synthetic as well as camera-captured images.
While using synthetic images allows highly controlled experiments, the training and evaluation based on camera-captured images demonstrates that our approach can be applied to real-world scenarios.

For creating synthetic images, we used the ShapeNet dataset \cite{Chang2015} of categorized 3D meshes that has also been used for many other 3D reconstruction tasks.
We rendered the 3D meshes using Blender with a resolution of typically $128\times128$ pixels and from multiple directions by using lighting hyperparameters different from the lighting hyperparameters that we used in the $SR$ of the RAN. 
This avoids unintended implicit supervision of the process.

In the general case we used sets of images from four azimuths ($\Delta_{\mathrm{azimuth}}=90^\circ$) for our training.
The results for this setting are shown in Fig. \ref{fig:results-classic}.
In this case, the viewpoints are from altitudes which are not contained in the training data, e.g., the training images of cars were always from an altitude of $0^\circ$---thus, a perfect reconstruction of a diagonal view of the models is harder.
Since our smooth renderer does not consider shadows, reconstructing the sofa in line 5 is especially hard.

In addition, we conducted studies on modified settings as presented and described in Fig.~\ref{fig:results-exp}.
In (a--c), we performed four-view trainings with the following modifications to the training data: In (a), we randomized the azimuth of the images with a standard deviation of $5^\circ$. 
In subfigure (b), we randomly assigned the position of the light source for each set of images. 
In subfigures (c1) and (c2), we trained on the \textit{car} and \textit{airplane} classes simultaneously.

In (d) and (e), we randomized, but supervised, the difference between azimuths for the four images. I.e., if multiple images have the same azimuth, the input data is effectively three or fewer images. 
In (d1) and (d2), we predicted the images from only two input images; in (e1--e4), we show single-view reconstructions. Since these reconstructions are trained on a single resp.~dual view only, the quality of entirely unseen parts of the reconstruction is lower.

Training itself was performed on up to 4 Nvidia V100, GTX Titan Xp, and GTX 1080 Ti GPUs on the basis of Float32. 
To enable a stable training, we already needed to clamp extreme values, esp.~for logarithmic and exponential functions, which was much worse when using Float16. 
While Float64 would be beneficial from a numerical perspective since it would allow to reduce/remove clamping, we continued with Float32 since Float64 is not efficiently supported by most GPUs, nowadays.

In our experiments, we used a uniform sphere with $162$ vertices and $320$ faces as base model.
When using an image size of $128\times128$ and a resolution of $320$ faces, our network has a memory footprint of almost 2GB per image in a batch, i.e., for a four-view training with batch size 1, we require $6.5$GB of VRAM; one iteration took 4 seconds on a single GTX Titan Xp.
When performing training on two GTX Titan Xp GPUs with batch size 2, the iteration time was $5.2$ seconds.
Training on a i9-7920X CPU @ 2.90GHz with 12 threads took $2$ minutes per iteration. %
To achieve faster training and lower memory footprint, we used image resolutions of $64\times64$ pixels, which was about four times faster and smaller.

When using huge batch sizes, we could not significantly increase the learning rate since high learning rates reduce the stability of the RAN. That is, because training with respect to competing alternating losses requires a low learning rate. The networks for the presented results have been trained for between one and three weeks on a single GTX Titan Xp.
 
When processing low resolution images ($64\times64$ pixels) combined with a high mesh resolution (642 vertices, $1280$ faces) faces have sub-pixel size. 
Thus, it occurs that single vertices dissociate themselves from the mesh since they are not visible any more.
The problem of dissociating vertices is even worse if, instead of considering the directed distance to all edges of the faces only the distance to the faces is considered in the smooth visibility test.
In the future we will alleviate this by going to higher image and network resolutions, which is possible when much more memory will be available on the GPU boards.

For training on camera-captured images, we used single-view images of shoes \cite{Yu2014}. Since these images are all typically taken from more or less the same direction, we use mirrored versions of the images and pretend this would be the view from the other side. We employed this small trick since many objects such as shoes are commonly roughly symmetric. Moreover, the back of the shoe could not be reconstructed without even having any training sample from the back side. Since this problem is highly ill-posed, our results could still be improved ---nevertheless, they are the first of their kind. Figure~\ref{fig:results-shoes} presents the results of this single-view 3D mesh reconstruction which was trained on camera-captured images from a single direction.

We also tried to use silhouette losses, given the silhouette of the object in the image is known, but found that such losses can both improve and worsen the results. In many cases, the accuracy drops because in the RANs internal representation the silhouette consistency does not hold---that is because there may be shifts or a small scaling factor between images of spaces $A$ and $B$.

\section{Discussion and Conclusion}\label{sec:discon}

In this work, we have demonstrated a robust way to reconstruct 3D geometry from only 2D data, eliminating the need for ground truth 3D models, or prior knowledge regarding materials and lighting conditions.
In addition, we have demonstrated how a globally differentiable renderer is crucial to the learning process---even if designing one induces differences in the appearance of the produced renderings. 
We alleviate this difference through the use of image domain translation.
The success of the reconstruction is driven by a restriction of the information flow and by the laziness of \texttt{pix2pix} networks, which easily perform image-style exchanges but struggle in changing the content of an image---a property that we exploit.
Thus, our approach is not informed by data but instead by an understanding of the real world.

In addition, we believe that our RAN architecture is suitable to more than just its current application of 3D reconstruction, but rather to a variety of inverse problems.
To do so, one could replace the smooth renderer with another smooth forward simulator to reconstruct the inverse of the respective algorithm without supervision.
In such a case, we would also distinguish between spaces $A$ and $B$ since the output of the smooth simulator is generally different from the output of its discrete counterpart.
Depending on the problem, it can be useful to shortcut the smooth algorithm to let the discriminator also consider the reconstructed space.

This novel architecture potentially leads to many interesting lines of future investigation.
The immediate direction would be training the reconstruction from single images.
The problem of training 3D reconstructions using a single view approach is, of course, the fact that the occluded side of the model is unknown.
We have overcome this issue by initially using training data from multiple viewpoints to familiarize the network with objects of a specific class, and then gradually switching to training the system to infer shapes using only a single image.
Another approach would be to use a differentiable renderer to create training images by supervising 3D shapes, materials and light settings.
This approach must be handled with care---while this is possible for synthetic settings, where the images can be rendered using an equivalent renderer, this is not possible when working with camera-captured images.
In addition, given the right training dataset, we believe the performance of the RAN for camera-captured single-view image 3D reconstructions could be significantly improved.

Another interesting path would be to allow decoding information like textures in the internal smooth image space. This could be done through adding (non-realistic) colors to the smooth renderer which would increase the representative power of the mentioned internal image space.
It would also be interesting to extend the developed renderer further to support also global illumination, which would enable the reconstructions of whole scenes.

Finally, we hope to inspire the research community to explore the full power of our smooth forward simulator and RAN architecture. We believe that it can be used for many other research objectives, such as the inverse-problem of iterated function systems or an unsupervised speech-to-text translation. The source code of our work will be publicly available.

\printbibliography

\newpage
\newpage
\begin{appendices}

\section{Implementation details}

\subsection{Losses}\label{apx:losses}

Our optimization of \texttt{p2v}, $a2b$, $b2a$, and $D$ involves adversarial losses, cycle-consistency losses, and mesh regularization losses.
Specifically, we solve the following optimization:
\begin{align*}
    \min_{\texttt{p2v}}\min_{a2b}\min_{b2a}\max_D \ 
    \mathcal{L}
\end{align*} 
or in greater detail
\begin{align*}
    \min_{\texttt{p2v}}\min_{a2b}\min_{b2a}\max_D \ 
    \sum_{i=1}^5\ \left(
    \alpha_i \cdot
    \mathcal{L}_i \right)
    + \mathcal{L}_{\mathrm{reg}}.
\end{align*} 
where $\alpha_i$ is a weight in $[0, 1]$, $\mathcal{L}$, $\mathcal{L}_i$, and $\mathcal{L}_{\mathrm{reg}}$ shall be defined below.

We define $b', b'' \in B$ and $a', a'' \in A$ in dependency of $a\in A$ according to Fig. \ref{fig:rancomb} as
\begin{align*}
    b' &= a2b(a)\\
    b'' &= SR\circ\texttt{p2v}(a)\\
    a' &= b2a(b')\\
    a'' &= b2a(b'')
\end{align*}
With that, our losses are 
\begin{align*}
    \mathcal{L}_{1} = 
    &\ \mathbb{E}_{a\sim A}[\log D(a, b'')] + \mathbb{E}_{a\sim A}[\log (1 - D(a, b'))] \\
    + &\ \mathbb{E}_{a\sim A}[\|b'' - b'\|_1] \\
    \mathcal{L}_{2} = 
    &\ \mathbb{E}_{a\sim A}[\log D(a, b'')] + \mathbb{E}_{a\sim A}[\log (1 - D(a'', b''))] \\
    + &\ \mathbb{E}_{a\sim A}[\|a'' - a\|_1] \\
    \mathcal{L}_{3} = 
    &\ \mathbb{E}_{a\sim A}[\log D(a, b')] + \mathbb{E}_{a\sim A}[\log (1 - D(a'', b'))] \\
    + &\ \mathbb{E}_{a\sim A}[\|a' - a\|_1] + \mathbb{E}_{a\sim A}[\|b'' - b'\|_1]\\
    \mathcal{L}_{4} = 
    &\ \mathbb{E}_{a\sim A}[\log D(a, b'')] + \mathbb{E}_{a\sim A}[\log (1 - D(a', b''))] \\
    + &\ \mathbb{E}_{a\sim A}[\|a' - a\|_1] + \mathbb{E}_{a\sim A}[\|b'' - b'\|_1] \\
    \mathcal{L}_{5} = 
    &\ \mathbb{E}_{a\sim A}[\log D(a, b')] + \mathbb{E}_{a\sim A}[\log (1 - D(a', b'))] \\
    + &\ \mathbb{E}_{a\sim A}[\|a' - a\|_1]
    .
\end{align*}

This results in our combined loss of (without weights)
\begin{align*}
    \mathcal{L} = 
    &\ \mathbb{E}_{a\sim A}[\log D(a, b')] \\
    + &\ \mathbb{E}_{a\sim A}[\log D(a, b'')] \\
    + &\ \mathbb{E}_{a\sim A}[\log (1 - D(a, b'))] \\
    + &\ \mathbb{E}_{a\sim A}[\log (1 - D(a', b'))] \\
    + &\ \mathbb{E}_{a\sim A}[\log (1 - D(a', b''))] \\
    + &\ \mathbb{E}_{a\sim A}[\log (1 - D(a'', b'))] \\
    + &\ \mathbb{E}_{a\sim A}[\log (1 - D(a'', b''))] \\
    + &\ \mathbb{E}_{a\sim A}[\|b'' - b'\|_1] \\
    + &\ \mathbb{E}_{a\sim A}[\|a' - a\|_1] \\
    + &\ \mathbb{E}_{a\sim A}[\|a'' - a\|_1] \\
    + &\ \mathcal{L}_{\mathrm{reg}}
    .
\end{align*}

$\mathcal{L}_{\mathrm{reg}}$ are the regularization losses on the reconstructed meshes.
These regularizations are presented with descending relevance:
\explainedstep{The angle of normals} of adjacent faces should be as similar as possible (loss uses the $L^2$ norm).
\explainedstep{The lengths of edges} should be as similar as possible (loss uses the $L^1$ norm).
\explainedstep{The distance to the mean vertex} of adjacent vertices should be as small as possible to imply a regular mesh and also reduce the curvature of the mesh (loss uses the $L^1$ norm).

\subsubsection{Training procedure}

We alternately train the different sections of our network in the following order:
\begin{enumerate}
    \item The discriminator $D$
    \item The translation from $B$ to $A$ ($b2a$)
    \item The components that perform a translation from $A$ to $B$ (\texttt{p2v}+SR, $a2b$)
\end{enumerate}
For each of these sections, we separately train the five losses $\mathcal{L}_i$.
For that, we used the Adam optimizer \cite{KingmaB14} and started the training with a learning rate of $10^{-6}$.

\subsection{Network Architectures}

Here, we describe the topologies of the components \texttt{p2v}, $a2b$, $b2a$, and $D$.
In our experiments we typically used an image resolution of $n\times n=128\times128$ and a number of vertices $v=162$---we will base the following details on that assumption.

Let \texttt{Ck} denote a Convolution--LeakyReLU layer with \text{k} filters of size $4\times4$ and a stride of $2$.
Let the negative slope of the LeakyReLU be $0.2$.
From the fifth convolutional layer on, we apply a $50\%$ dropout.

The \textbf{\texttt{pix2pix}} network is a symmetric residual network with the following blocks defining the first half:
\texttt{C64-C128-C256-C512-C512-C512-C512}

\textbf{\texttt{pix2vex}} is based on the \texttt{pix2pix} network.
It is followed by two fully connected layers ($n^2\rightarrow \left\lfloor\frac{n^2+3\cdot v}{2}\right\rfloor$ and $\left\lfloor\frac{n^2+3\cdot v}{2}\right\rfloor \rightarrow 3\cdot v$) and the sigmoid function.
The remainder of the \texttt{pix2vex} architecture is explained in Section \ref{sec:pix2vex} in greater detail.

\textbf{\textit{a2b}} and \textbf{\textit{b2a}} are \texttt{pix2pix} networks with the first two or more residual layers.
This is followed by the sigmoid function.

The \textbf{discriminator \textit{D}} is defined as
\texttt{C64-}\texttt{C128-} \texttt{C256-}\texttt{C512-}\texttt{C1}
followed by the sigmoid function.

\subsection{Stabilization of the smooth Z-buffer}

A problem of the in Section \ref{sub:smoothzbuffer} described smooth z-buffer is that pixels that are not covered by any triangle can appear as if they had visible triangles.
That is because, if all weights are very low, even weights close to zero can cause a high value in the weighted SoftMin.
To stabilize this, we add two triangles covering the background and reduce the visibility value with increasing distance from triangles.

 \end{appendices}

\end{document}